\documentclass[10pt,journal,final,finalsubmission,onecolumn]{IEEEtran}

\usepackage{graphicx}  
\begin{document}
\title{Calibration of an Articulated Camera System with Scale Factor Estimation}

\author{CHEN~Junzhou$^*$, Kin~Hong~WONG \thanks{$^*$ Corresponding Author} \thanks{J. Chen is
with the School of Information Science \& Technology Southwest Jiaotong University, China. ~\emph{E-mail
address:} jzchen@swjtu.edu.cn.}
\thanks{K. H. Wong is with the Department of Computer Science and
Engineering, the Chinese University of Hong Kong, Shatin, NT, Hong
Kong. ~\emph{E-mail address:} khwong@cse.cuhk.edu.hk.}
\thanks{This work is supported by the National Natural Science Foundation of China (No.61003143).}
}

\maketitle \thispagestyle{empty}
\begin{abstract}
Multiple Camera Systems (MCS) have been widely used in many vision
applications and attracted much attention recently. There are two
principle types of MCS, one is the Rigid Multiple Camera System
(RMCS); the other is the Articulated Camera System (ACS). In a RMCS,
the relative poses (relative 3-D position and orientation) between
the cameras are invariant. While, in an ACS, the cameras are
articulated through movable joints, the relative pose between them
may change. Therefore, through calibration of an ACS we want to find
not only the relative poses between the cameras but also the
positions of the joints in the ACS.

Although calibration methods for RMCS have been extensively
developed during the past decades, the studies of ACS calibration
are still rare. In this paper, we developed calibration algorithms
for the ACS using a simple constraint: the joint is fixed relative
to the cameras connected with it during the transformations of the
ACS. When the transformations of the cameras in an ACS can be
estimated relative to the same coordinate system, the positions of
the joints in the ACS can be calculated by solving linear equations.
However, in a non-overlapping view ACS, only the ego-transformations
of the cameras and can be estimated. We proposed a two-steps method
to deal with this problem. In both methods, the ACS is assumed to
have performed general transformations in a static environment. The
efficiency and robustness of the proposed methods are tested by
simulation and real experiments. In the real experiment, the
intrinsic and extrinsic parameters of the ACS are obtained
simultaneously by our calibration procedure using the same image
sequences, no extra data capturing step is required. The
corresponding trajectory is recovered and illustrated using the
calibration results of the ACS. Since the estimated translations of
different cameras in an ACS may scaled by different scale factors, a
scale factor estimation algorithm is also proposed. To our
knowledge, we are the first to study the calibration of ACS.
\end{abstract}
\section{Introduction}\label{sec:introduction}
Calibration of a Multiple Camera System (MCS) is an essential step
in many computer vision tasks such as SLAM (Simultaneous
Localization and Map), surveillance, stereo and metrology
\cite{kaess2006vsm,Baker2004Argus,Dockstader2001Tracking,furukawa2009accurate,garcia2013geometric,morais2013multiple}.
Both the intrinsic and extrinsic parameters of the MCS are required
to be estimated before the MCS can be used. The intrinsic parameters
\cite{heikkila1997fsc,Hartley2004} describe the internal camera
geometric and optical characteristics of each camera in the MCS. In
a Rigid Multiple Camera System (RMCS), the cameras are fixed to each
other. The extrinsic parameters \cite{caprile1990uvp} of a RMCS
describe the relative pose (the relative 3-D position and
orientation, totally, six degrees of freedom) between the cameras in
the MCS. Calibration methods of the intrinsic parameters of a camera
are well established \cite{shah1996ipc,zhang00flexible}. Calibration
methods for the extrinsic parameters of a RMCS are also widely
studied. For instance, Maas proposed an automatic RMCS calibration
technique with a moving reference bar which can be seen by all
cameras \cite{Maas1998M-Calibration}. Antone and Teller developed an
algorithm which recovers the relative poses of cameras by
overlapping portions of the outdoor scene
\cite{AntoneTeller2002sec}. Baker and Aloimonos presented RMCS
calibration methods using calibration objects such as a wand with
LEDs or a rigid board with known patterns
\cite{Baker2000calibration,Baker2003calibration}. Dornaika proposed
a stereo rig self-calibration method by the monocular epipolar
geometries and geometric constraints of a moving RMCS, in which only
the feature correspondences between the monocular images of each
camera are required \cite{dornaika2007scs}.
\begin{figure}[t]
  \centering
  \includegraphics[width=2.7in]{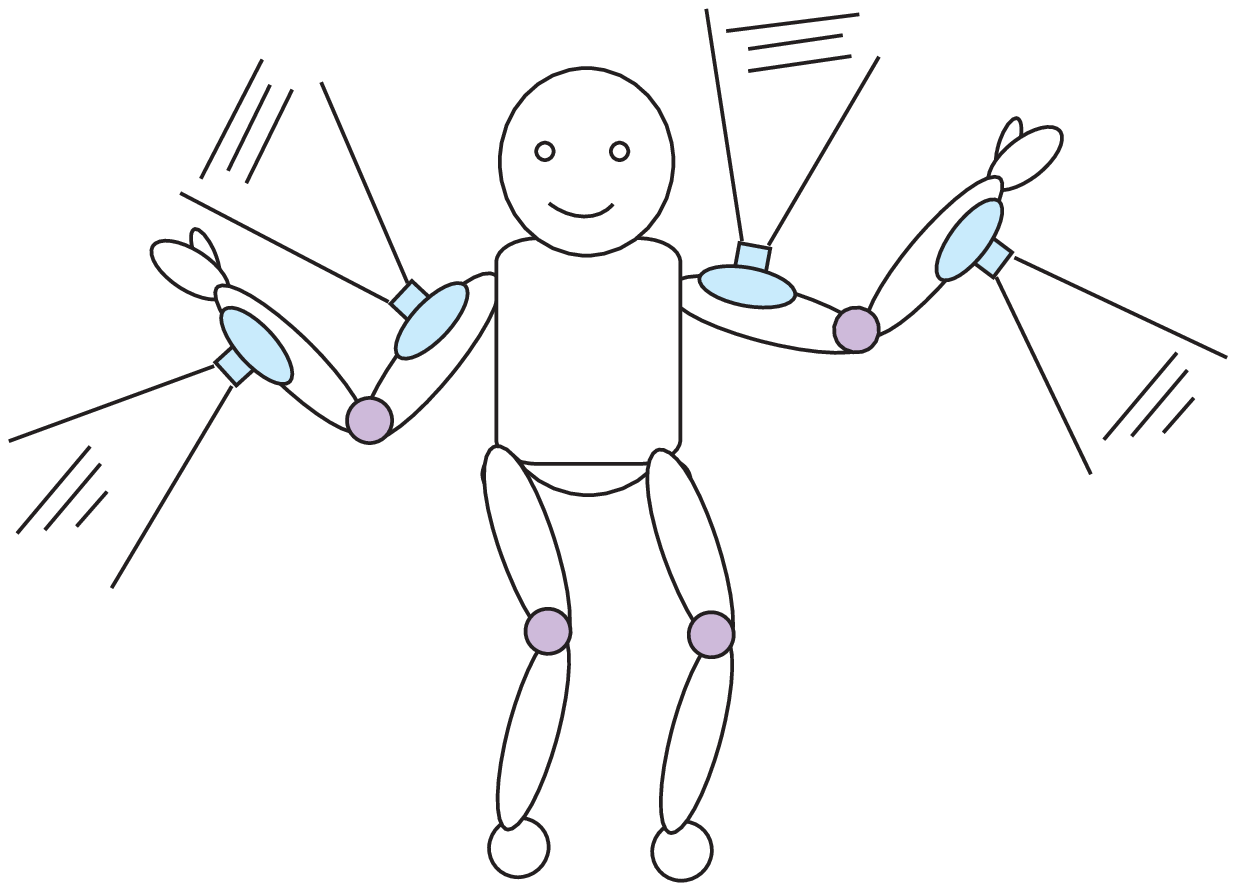}
  \caption{A Robot with Four Cameras Attached on It, Where the Cameras are Articulated.}\label{fig:robot}
\end{figure}
In hand-eye calibration, it is demonstrated that when a sensor is
mounted on a moving robot hand, the relationship between the sensor
coordinate system and hand coordinate system can be calculated by
the motion information of the hand and the sensor
\cite{tsai1989ntf,horaud1995hec,malti2012hand}. One example of using kinematic
information of the cameras for RMCS is discussed by Caspi and Irani
\cite{caspi2002ano}, they indicated that if the cameras of a
non-overlapping view RMCS are close to each other and share a same
projection center, their recorded image sequences can be aligned
effectively by the estimated transformations inside each image
sequence.

However, in some types of MCS, the relative poses between the
cameras are not fixed, hence the calibration methods for the RMCS
cannot be used directly. In Figure \ref{fig:robot}, a novel
application of limb pose estimation by attaching cameras on the arms
of a robot is shown. On each arm of the robot, two cameras are
articulated to each other through the elbow joint of the arm. When
the robot moves, the relative pose between the cameras may change,
while, the coordinate of the elbow joint relative to each camera
attached on the corresponding arm is invariant. In this paper, such
a type of MCS is named as Articulated Camera System (ACS). The joint
of the elbow is named as the \emph{joint} in the ACS.

ACSs can be easily found in the real world, such as camera systems
attached on human, robots and animals. Before using an ACS, it has
to be calibrated. However, there are still some unsolved problems:
(i) In an ACS with overlapping view, traditional calibration methods
cannot estimate the positions of the joints in the ACS. (ii) In a
non-overlapping view ACS, neither the positions of the joints in the
ACS nor the relative poses between the cameras in the ACS can be
estimated by traditional calibration methods.

These considerations in mind motivate us to develop the technologies
in this paper. The rest of this paper are organized as follows:
Section \ref{sec:overlap} and \ref{sec:non_overlap} analysis the
constraints in a moving ACS. The corresponding calibration methods
are proposed. Section \ref{sec:simulation} and \ref{sec:real}
evaluate the proposed method by simulation and real experiment. In
section \ref{sec:conclusion}, a brief conclusion and the future plan
are presented.

\section{Calibration of ACS with Overlapping Views}
\label{sec:overlap}
\begin{figure}[h]
 \centering
 \includegraphics[width=2.7in]{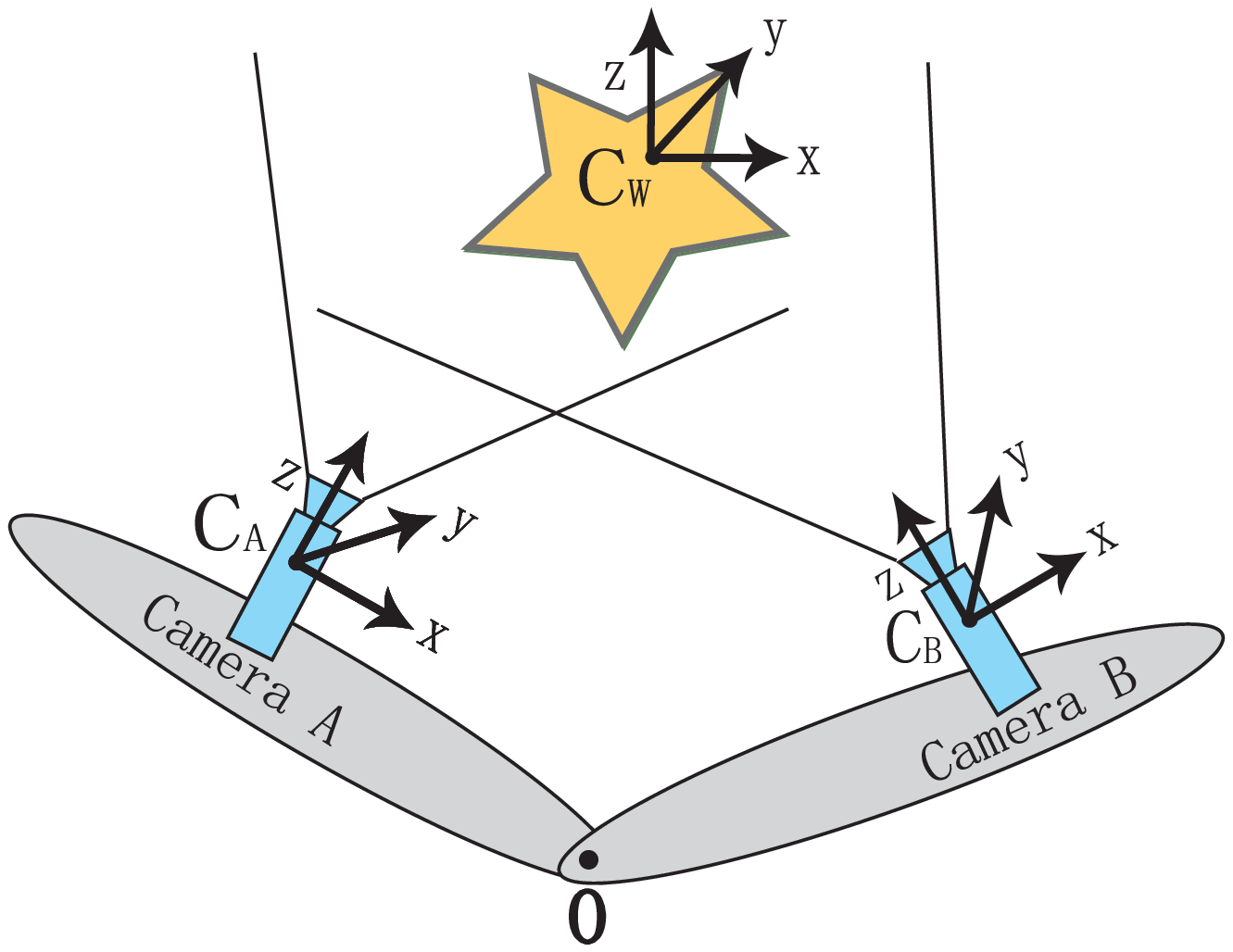}\\
 \caption{An Articulated Camera System with Overlapping Views}\label{fig:articulate}
\end{figure}
Suppose two rigid objects are articulated at joint O and two cameras
(camera A and B) are fixed on the two rigid objects respectively
(See Figure \ref{fig:articulate}). Let $C_A$ be the coordinate
system of camera A, $C_B$ the coordinate system of camera B. Suppose
there are enough feature correspondences between the cameras so that
the pose of $C_A$ and $C_B$ referring to the same coordinate system
$C_W$ can be estimated. Therefore, the relative pose between $C_A$
and $C_B$ is known. We want to find the position of O in the ACS.
Let $\mathbf{H}_{AW}$ and $\mathbf{H}_{BW}$ be the Euclidean
transformation matrixes describe the $C_A$ and $C_B$ relative to
$C_W$, so that for any point $P$:
\begin{equation}\label{eq:h_aw}
P_A = \mathbf{H}_{AW} P_W = \left[\begin{array}{c c}
                           \mathbf{R}_{AW} & T_{AW}\\
                           0 & 1 \end{array}\right]
                           \left[\begin{array}{c}\bar{P}_W\\1\end{array}\right]
\end{equation}
\begin{equation}\label{eq:h_bw}
P_B = \mathbf{H}_{BW} P_W = \left[\begin{array}{c c}
                           \mathbf{R}_{BW} & T_{BW}\\
                           0 & 1
                         \end{array}\right]
                         \left[\begin{array}{c}\bar{P}_W\\1\end{array}\right]
\end{equation}
, where $\mathbf{R}$ is the $3\times 3$ rotation matrix, $T$ is a
$3\times1$ vector, $P_W$, $P_A$ and $P_B$ are the homogenous
coordinates of the 3-D Point $P$ relative to $C_W$, $C_A$ and $C_B$
respectively, $\bar{P}$ is a $3\times1$ vector.

According to equations (\ref{eq:h_aw}) and (\ref{eq:h_bw}):
\begin{equation}
P_W = \mathbf{H}_{AW}^{-1} P_A = \mathbf{H}_{BW}^{-1} P_B
\end{equation}
\begin{equation}
\mathbf{H}_{AW}^{-1}P_A  - \mathbf{H}_{BW}^{-1}P_B = 0
\end{equation}
\begin{equation}
\left[\begin{array}{c c}
\mathbf{R}_{AW}^T & -\mathbf{R}_{AW}^T T_{AW}\\
0 & 1
\end{array}\right] \left[\begin{array}{c}
        \bar{P}_A \\
        1
      \end{array}\right] -  \left[\begin{array}{c c}
\mathbf{R}_{BW}^T & -\mathbf{R}_{BW}^T T_{BW}\\
0 & 1
\end{array}\right] \left[\begin{array}{c}
        \bar{P}_B \\
        1
      \end{array}\right] = 0
\end{equation}
\begin{equation}\label{eq:papb}
\mathbf{R}_{AW}^T \bar{P}_A - \mathbf{R}_{BW}^T \bar{P}_B =
\mathbf{R}_{AW}^T T_{AW} - \mathbf{R}_{BW}^TT_{BW}
\end{equation}
, where $\mathbf{R}^T$ is the transpose of $\mathbf{R}$. Suppose the
ACS performed $n$ transformations. Let $\mathbf{H}_{AW}^i$ and
$\mathbf{H}_{BW}^i$ be the Euclidean transformation matrixes
describe the $C_A$ and $C_B$ relative to $C_W$ after the $i$-th
transformation of the ACS. According to equation (\ref{eq:papb}):
\begin{equation}\label{eq:Oab}
 (\mathbf{R}_{AW}^i)^T \bar{P}_A - (\mathbf{R}_{BW}^i)^T \bar{P}_B = (\mathbf{R}_{AW}^i)^T T_{AW}^i - (\mathbf{R}_{BW}^i)^T
 T_{BW}^i
\end{equation}

Let $\tilde{O} = \left[\begin{array}{cc} \bar{O}_A^T & \bar{O}_B^T
\end{array}\right]^T$, where $\bar{O}_A$ and $\bar{O}_B$ are the coordinates of the joint O relative to $C_A$ and $C_B$
respectively. Equation (\ref{eq:Oab}) can be rewritten as:
\begin{equation}\label{eq:O}
\left[\begin{array}{cc} (\mathbf{R}_{AW}^i)^T &
-(\mathbf{R}_{BW}^i)^T\end{array}\right]
                         \tilde{O} =
                         (\mathbf{R}_{AW}^i)^T T_{AW}^i - (\mathbf{R}_{BW}^i)^T T_{BW}^i
\end{equation}

Since camera A and B are fixed on the articulated rigid objects,
$\tilde{O}$ is invariant during the transformation of the ACS. The
transformations ($\mathbf{R}_{AW}^i$, $\mathbf{R}_{BW}^i$,
$T_{AW}^i$ and $T_{BW}^i$ for $i \in [1 \dots n]$) of the camera
coordinate systems are calculated by the projected image sequences.
We propose that $\tilde{O}$ can be estimated by a least squares
method, when the ACS has moved to many different positions and
captured enough samples of
$\mathbf{R}_{AW}^i$, $\mathbf{R}_{BW}^i$, $T_{AW}^i$ and $T_{BW}^i$.\\

The above derivation shows that although the location of the joint
$O_W^i$ in world coordinates is not constant, it equals
$(H_{AW}^i)^{-1}O_A$ or $(H_{BW}^i)^{-1}O_B$  because the cameras
can not move completely independent as they are connected with a
joint. The joint location can be calculated by the 1D subspace
intersection of the camera transformation matrices.

\section{Calibration of Non-Overlapping View ACS}
\label{sec:non_overlap}

\begin{figure}[h]
\centering
  \includegraphics[width=3in]{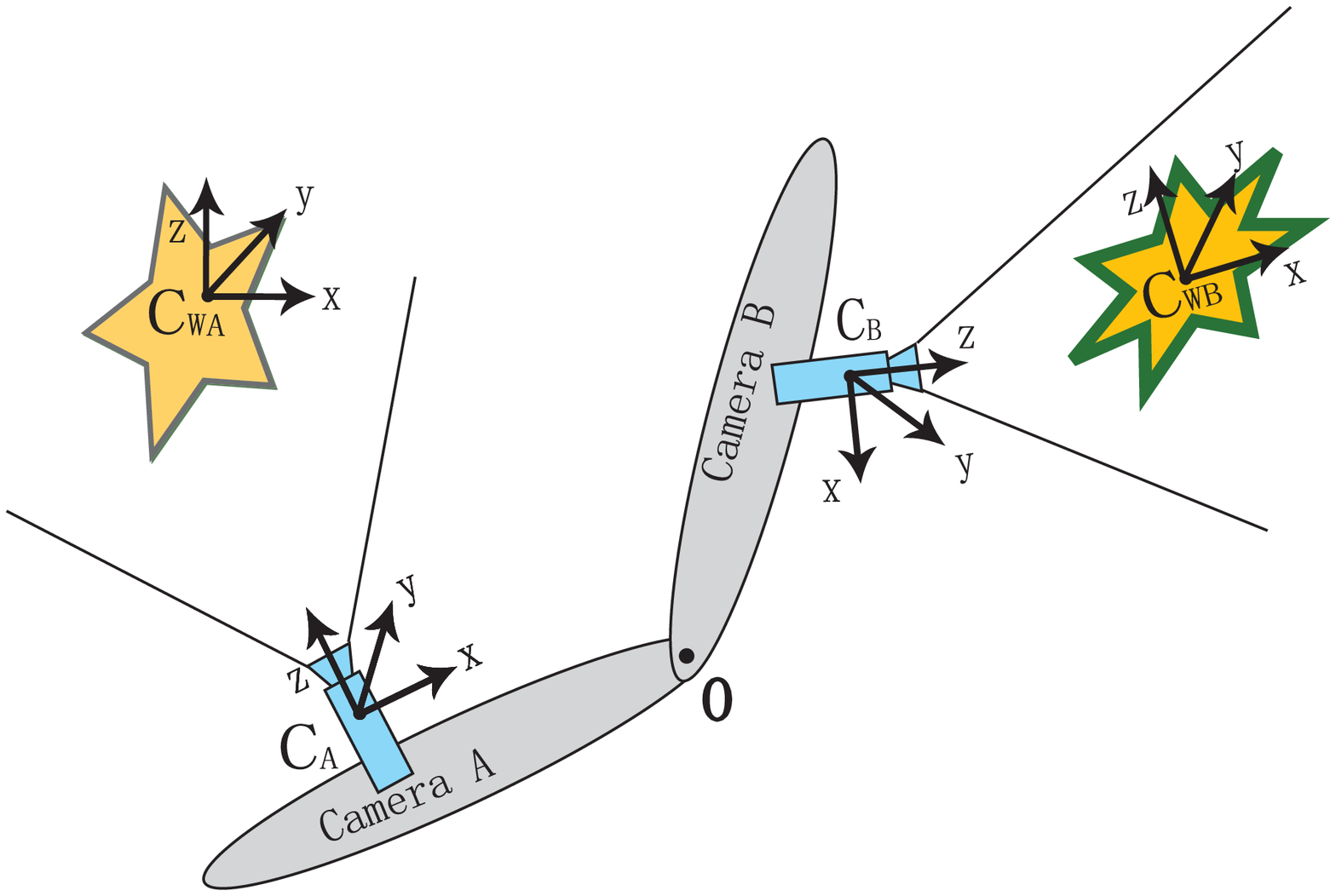}
  \caption{A Non-overlapping View Articulated Camera System}
  \label{fig:non_overlap}
\end{figure}

In many situations, there is no overlapping view between the cameras
in an ACS. And the lack of common features makes the calibration
method proposed in section \ref{sec:overlap} become invalid (See
Figure \ref{fig:non_overlap}). Moreover, since the relative pose
between the cameras in the ACS cannot be estimated by the
overlapping views, the calibration of the relative poses between the
non-overlapping view cameras is also required. In this section, a
calibration method based on the ego-motion information of the
cameras in an ACS is discussed.
\subsection{Recovering the Position of the Joint Relative to the Cameras in the
ACS}\label{sec:special_motion}

Let $C_{A}^{init}$ and $C_{B}^{init}$ be the coordinate systems of
camera A and B respectively at the initial state (time $t = 0$).
Suppose the ACS performs $n$ transformations. Since the coordinate
of the joint O relative to camera A is fixed during the
transformation of the ACS. At time $t = i$, we have:
\begin{equation}
\label{eq:OAi} O_A^i = \mathbf{H}_A^i O_A = \left[
{{\begin{array}{cc}
 \mathbf{R}_A^i& T_A^i \\
 {0}  & {1} \\
\end{array} }} \right]{O}_{A}
\end{equation}
, where $\mathbf{H}_A^i$ is the Euclidean transformation matrix of
camera A at time $i$ relative to $C_A^{init}$. $\mathbf{R}_A^i$ and
$T_A^i$ describe the orientation and origin of camera A at time $i$
relative to $C_A^{init}$. Also $O_A$ is the coordinate of point O at
initial state relative to $C_A^{init}$, and $O_A^i$ is the
coordinate of point O at time $i$ relative to $C_A^{init}$.

If the position of the joint O relative to $C_A^{init}$ is fixed
during the transformations of the ACS, we have: $O_A^i = O_A$,
$\forall i \in [1, \dots, n]$. For $i$-th transformation of the ACS,
according to equation (\ref{eq:OAi}):

\begin{equation}
\label{eq:OA_fixed} O_A = \mathbf{H}_A^i O_A = \left[
{{\begin{array}{cc}
 \mathbf{R}_A^i& T_A^i \\
 {0}  & {1} \\
\end{array} }} \right]{O}_{A}
\end{equation}
\begin{equation}
\label{eq:OA_fixed_I} (\mathbf{R}_A^i -I) \bar{O}_A  =  -T_A^i
\end{equation}

Let $\mathbf{M}_A = [(\mathbf{R}_A^1 -I)^T, (\mathbf{R}_A^2 -I)^T,
\dots, (\mathbf{\mathbf{R}}_A^n -I)^T]^T$, $\tilde{T}_A =
[(T_A^1)^T, (T_A^2)^T, \dots, (T_A^n)^T]^T$, we have:
\begin{equation}\label{eq:fixed_o}
  \mathbf{M}_A \bar{O}_A = -\tilde{T}_A
\end{equation}

Since the transformations ($\mathbf{R}_{A}^i$ and $T_{A}^i$,
$\forall i \in [1 \dots n]$) of camera A can be calculated by the
projected image sequence. We propose $\bar{O}_A$ can be estimated by
a least squares method. Similarly, $\bar{O}_B$ can also be
estimated. Therefore, $O_A$ and $O_B$ are recovered.

\subsection{The Uniqueness of the Joint Pose Estimation}
If the different segments of the articulated camera system (ACS) are
connected by 1D rotational joints (connected by point rotational
joints) and the ACS can perform general transformations, the
solution of the joint pose estimation is unique:

For the joint pose estimation method using special motion (in
section \ref{sec:special_motion}). Suppose the solution of the joint
pose estimation is not unique, there must exist at least two
different 3D points $\bar{O}_1$ and $\bar{O}_2$ satisfy equation
(\ref{eq:fixed_o}).  We have: $\mathbf{M}_A  \bar{O}_1 =
-\tilde{T}_A$ and $ \mathbf{M}_A \bar{O}_2 = -\tilde{T}_A$.
Therefore, any point $\bar{P} = s \bar{O}_1 + (1-s) \bar{O}_2$ will
also satisfy equation (\ref{eq:fixed_o}), where $s$ is an arbitrary
scalar. According to the definition of $\bar{P}$, $\bar{P}$ is the
point on the line passing through the points $\bar{O}_1$ and
$\bar{O}_2$. Since $\bar{P}$ satisfy equation (\ref{eq:fixed_o})
represents that the position of the point $P$ relative to the camera
in the ACS is invariant during the transformation of the ACS, it
means the different segments of ACS are connected by the 2D
rotational axis instead of the 1D rotational joints. The position of
the points on the 2D rotational axis relative to the camera in the
ACS is invariant during the transformation of the ACS. However, it
conflicts with the assumption. Similarly, the uniqueness of the
joint pose estimation method using overlapping views (in section
\ref{sec:overlap}) can also be verified.

\subsection{Recovering the Relative Pose Between the Cameras of the Non-overlapping view ACS} \label{sec:recover_pose}
Let $\mathbf{H}_{BA}$ be the Euclidean transformation matrix between
$C_{A}^{init}$ and $C_{B}^{init}$, so that for any point $P$:

\begin{equation}
\label{eq:BA_relation} {P}_{B} =\mathbf{H}_{BA} P_{A} = \left[
{{\begin{array}{cc}
 {\mathbf{R}_{BA}}  & T_{BA}\\
 {0}  & {1}  \\
\end{array} }} \right]{P}_{A} =\mathbf{H}_{BA}P_{A}
\end{equation}
, where $P_A$ and $P_B$ are the homogenous coordinate of Point $P$
relative to $C_{A}^{init}$ and $C_{B}^{init}$ respectively.\\

The relative pose ($\tilde{\mathbf{R}}_{BA}$ and $\tilde{T}_{BA}$)
between $C_{A}^{init}$ and $C_{B}^{init}$ is defined as:
\begin{equation}\label{eq:R_define}
\tilde{\mathbf{R}}_{BA} = {\mathbf{R}}_{BA}^T
\end{equation}
\begin{equation}\label{eq:T_define}
\tilde{T}_{BA} = -{\mathbf{R}}_{BA}^T {T}_{BA}
\end{equation}

Let $O_B^i$ be the coordinate of joint O at time $i$ relative to
$C_B^{init}$. Since the coordinate of the joint O  relative to
camera B is invariant:
\begin{eqnarray}
\label{eq:OB1} O_B^i &=&\left[ {{\begin{array}{cc}
 {\mathbf{R}_B^i }  & {T_B^i }  \\
 {0}  & {1}  \\
\end{array} }} \right]O_B \nonumber\\
&=&\left[ {{\begin{array}{cc}
 {\mathbf{R}_B^i }  & {T_B^i }  \\
 {0}  & {1}  \\
\end{array} }} \right]\left[ {{\begin{array}{cc}
 \mathbf{R}_{BA}  & T_{BA}  \\
 {0}  & {1}  \\
\end{array} }} \right]O_{A} \nonumber\\&=&\left[ {{\begin{array}{cc}
 {\mathbf{R}_B^i \mathbf{R}_{BA}}  &
\mathbf{R}_B^i T_{BA} + T_B^i \\
 {0}  & {1}  \\
\end{array} }} \right]O_{A}
\end{eqnarray}

According to equations (\ref{eq:OAi}) and (\ref{eq:BA_relation}):
\begin{eqnarray}
\label{eq:OB1_AB} O_B^i &=& \mathbf{H}_{BA}O_A^i
\nonumber\\&=&\left[\begin{array}{cc}
 \mathbf{R}_{BA} & T_{BA} \\
 {0}  & {1}  \\
\end{array} \right] \left[ \begin{array}{cc}
 \mathbf{R}_A^i & T_A^i\\
 {0}  & {1}  \\
\end{array}  \right]O_{A} \nonumber\\
&=&\left[ {{\begin{array}{cc}
 \mathbf{R}_{BA} {\mathbf{R}}_A^i  &
{\mathbf{R}_{BA}T_A^i + {T_{BA}}}\\
 {0}  & {1}\\
\end{array} }} \right]{O}_{A}
\end{eqnarray}

According to equations (\ref{eq:OB1}) and (\ref{eq:OB1_AB}):
\begin{equation}
\left[{{\begin{array}{cc}
 {\mathbf{R}_B^i \mathbf{R}_{BA}}  & {\mathbf{R}_B^i T_{BA} + T_B^i}  \\
 {0}  & {1}  \\
\end{array} }} \right]\left[ {{\begin{array}{c}
 \bar{O}_A  \\
 {1}  \\
\end{array} }} \right]
= \left[ {{\begin{array}{cc}
 {\mathbf{R}_{BA}\mathbf{R}_A^i }  & \mathbf{\mathbf{R}}_{BA}T_A^i + {T_{BA}}\\
 {0}  & {1}  \\
\end{array} }} \right]\left[ {{\begin{array}{c}
 \bar{O}_A  \\
 {1}  \\
\end{array} }} \right]
\end{equation}
\begin{equation}
\left[ {{\begin{array}{cc}
 {\mathbf{R}_B^i \mathbf{R}_{BA}\bar{O}_A
+ \mathbf{R}_B^i
T_{BA} + T_B^i}  \\
 {1}  \\
\end{array} }} \right] =\left[ {{\begin{array}{cc}
 {\mathbf{R}_{BA}\mathbf{R}_A^i \bar{O}_A
+\mathbf{R}_{BA}T_A^i + T_{BA}}
\\
 {1}  \\
\end{array} }} \right]
\end{equation}

\begin{equation}\label{eq:cst_non_overlap}
\mathbf{R}_B^i \mathbf{R}_{BA}\bar {O}_A + \mathbf{R}_B^i T_{BA} -
 \mathbf{R}_{BA}\mathbf{R}_A^i \bar{O}_A
-\mathbf{R}_{BA}T_A^i+ T_B^i - T_{BA} = 0
\end{equation}

Since $\bar{O}_A$ can be estimated by the method discussed in
section \ref{sec:recover_pose}, the $\mathbf{R}_{BA}$ and $T_{BA}$
can be estimated by a least square method, when the ACS perform
enough general motions.

In our simulation and real experiment, the estimated $R_{BA}$ is
refined by a method discussed in \cite{zhengyou:ReportCalib}. Then
the roll, pitch and yaw corresponding to the $R_{BA}$ are estimated
according to the definition of the rotation matrix
\cite{Hartley2004}. Let $R_{BA}=M(r,p,y)$, where $r$ $p$ and $y$ are
the corresponding roll, pitch and yaw of $\mathbf{R}_{BA}$, $M$ is a
function from roll, pitch and yaw to the corresponding rotation
matrix. Then, the $r$, $p$, $y$, $T_{BA}$ and $\bar{O}_A$ are
optimized by minimizing the nonlinear error function:
\begin{eqnarray}\label{eq:opt}
  E(r,p,y,T_{BA},O_A) =\sum_{i=1}^n(\mathbf{R}_B^i M(r,p,y)\bar {O}_A + \mathbf{R}_B^i T_{BA}&& \nonumber\\
  - M(r,p,y) \mathbf{R}_A^i \bar{O}_A
-M(r,p,y)T_A^i+ T_B^i - T_{BA})&&
\end{eqnarray}
using a Levenberg-Marquardt method. Finally, the $R_{BA}$ is
recovered from the optimized $r$, $p$ and $y$. The relative pose
between the $C_{A}^{init}$ and $C_{B}^{init}$ is calculated by
equations (\ref{eq:R_define}) and (\ref{eq:T_define}).
\section{Dealing With Unknown Scale Factors} \label{sec:deal_scale}
The non-overlapping view ACS calibration method discussed above
depends on the ego-motion information of the cameras in the ACS.
However, if the model of the scene is unknown, the estimated
ego-translations of the cameras may be scaled by different unknown
scale factors. These unknown scale factors must be considered in the
extrinsic calibration process.\\

\subsection{Model Analysis}\label{sec:model_scale}
Let $T_{A}$ and $T_{B}$ be the true ego-translation of camera A and
B in the world coordinate system, $\hat{T}_{A}$ and $\hat{T}_{B}$ be
the estimated ego-translations of camera A and B found by an SFM
method, $\mu_A$ and $\mu_B$ be the corresponding unknown scale
factors. So that:
\begin{equation}\label{eq:mu_A}
\hat{T}_A = \mu_A T_{A}
\end{equation}
\begin{equation}\label{eq:mu_B}
\hat{T}_B = \mu_B T_{B}
\end{equation}

Let $\hat{\bar{O}}_A$ be the pose of the joint relative to $C_A$
calculated with the estimated motion. Equation (\ref{eq:OA_fixed_I})
can be rewritten as:

\begin{equation}\label{eq:scale_O_no_use}
(\mathbf{R}_A^i -I) \hat{\bar{O}}_A  =  -\hat{T}_A^i = -\mu_A T_A^i
\end{equation}
\begin{equation}\label{eq:scale_O}
(\mathbf{R}_A^i -I) \frac{\hat{\bar{O}}_A}{\mu_A}  = -T_A^i
\end{equation}

Compare equation (\ref{eq:scale_O}) with equation
(\ref{eq:OA_fixed_I}), we have:
\begin{equation} \label{eq:scale_O_mu}
\hat{\bar{O}}_A = \mu_A \bar{O}_A
\end{equation}

Let $\hat{\mathbf{R}}_{BA}$ and $\hat{T}_{BA}$ be the extrinsic
parameters calculated using the estimated motions and joint pose.
Equation (\ref{eq:cst_non_overlap}) can be rewritten as:

\begin{equation}\label{eq:scale_cst_non_overlap}
\mathbf{R}_B^i \hat{\mathbf{R}}_{BA}\hat{\bar {O}}_A +
\mathbf{R}_B^i \hat{T}_{BA} - \hat{\mathbf{R}}_{BA}\mathbf{R}_A^i
\hat{\bar{O}}_A - \hat{\mathbf{R}}_{BA}\hat{T}_A^i+ \hat{T}_B^i -
\hat{T}_{BA} = 0
\end{equation}

According to equation (\ref{eq:mu_A}), (\ref{eq:mu_B}),
(\ref{eq:scale_O_mu}) and (\ref{eq:scale_cst_non_overlap}):

\begin{equation} 
\mathbf{R}_B^i \hat{\mathbf{R}}_{BA}\mu_A\bar {O}_A + \mathbf{R}_B^i
\hat{T}_{BA} - \hat{\mathbf{R}}_{BA}\mathbf{R}_A^i \mu_A\bar{O}_A -
\hat{\mathbf{R}}_{BA}\mu_A{T}_A^i+ \mu_B{T}_B^i - \hat{T}_{BA} = 0
\end{equation}
\begin{equation}\label{eq:scale_cst_2}
\mathbf{R}_B^i \frac{\mu_A}{\mu_B}\hat{\mathbf{R}}_{BA}\bar {O}_A +
\mathbf{R}_B^i \frac{1}{\mu_B}\hat{T}_{BA} -
\frac{\mu_A}{\mu_B}\hat{\mathbf{R}}_{BA}\mathbf{R}_A^i \bar{O}_A -
\frac{\mu_A}{\mu_B}\hat{\mathbf{R}}_{BA}{T}_A^i+ {T}_B^i -
 \frac{1}{\mu_B}\hat{T}_{BA} = 0
\end{equation}

Let:
\begin{equation}\label{eq:factor_r} \frac{\mu_A}{\mu_B}
\hat{\mathbf{R}}_{BA} = \mathbf{\breve{R}}_{BA}
\end{equation}
\begin{equation}\label{eq:factor_t}
\frac{1}{\mu_B} \hat{T}_{BA} = \breve{T}_{BA}
\end{equation}

Equation (\ref{eq:scale_cst_2}) can be rewritten as:
\begin{equation}\label{eq:scale_cst_3}
\mathbf{R}_B^i \breve{\mathbf{R}}_{BA}\bar {O}_A + \mathbf{R}_B^i
\breve{T}_{BA} - \breve{\mathbf{R}}_{BA}\mathbf{R}_A^i \bar{O}_A -
\breve{\mathbf{R}}_{BA}{T}_A^i+ {T}_B^i - \breve{T}_{BA} = 0
\end{equation}

Since the equations (\ref{eq:scale_cst_3}) and
(\ref{eq:cst_non_overlap}) are exactly the same, we have:

\begin{equation}\label{eq:factor_R_pi}
\mathbf{R}_{BA} = \breve{\mathbf{R}}_{BA}
\end{equation}
\begin{equation}\label{eq:factor_T_pi}
T_{BA} =  \breve{T}_{BA}
\end{equation}

Therefore:

\begin{equation}\label{eq:factor_R_pi}
\hat{\mathbf{R}}_{BA} = \frac{\mu_B}{\mu_A}\mathbf{\breve{R}}_{BA} =
\phi_{BA}\mathbf{R}_{BA}
\end{equation}
\begin{equation}\label{eq:factor_T_pi}
\hat{T}_{BA} = {\mu_B} \breve{T}_{BA} = {\mu_B} T_{BA}
\end{equation}

Where $\phi_{BA}= \frac{\mu_B}{\mu_A}$. Equations
(\ref{eq:factor_R_pi}) and (\ref{eq:factor_T_pi}) show that the
estimated rotation matrix $\hat{\mathbf{R}}_{BA}$ will be scaled by
the relative scale factor (the ratio of the scale factors of the
cameras) and the estimated relative translation will be scaled by
the same scale factor of camera $B$. In the next section, we will
discuss the estimation of the relative scale factor.

\subsection{Rotation Matrix and Relative Scale Factor Estimation}\label{sec:rotation_refine}

Let $\mathbf{R}'= \phi \mathbf{R} + N$, where $\mathbf{R}$ is a $3
\times 3$ rotation matrix and $\mathbf{R}^T \mathbf{R} = I$, $\phi$
is an unknown scale factor, $N$ is a $3 \times 3$ unknown noise
matrix. We want to recover $\mathbf{R}$ and $\phi$ from
$\mathbf{R}'$. According to the definition, we have:
\begin{equation}
\mathbf{R}' = \phi \mathbf{R} + N = \phi(\mathbf{R} +
\frac{N}{\phi}) = \phi M
\end{equation}

Where $M = \mathbf{R} + \frac{N}{\phi}$.

Let the singular value decomposition of $M$ be $UDV^T$, where $D =
diag(\sigma _1, \sigma _2, \sigma _3)$. As
 illustrated in appendix C of \cite{zhengyou:ReportCalib}, $r$ can be
approximated by:

\begin{equation}\label{eq:app_rotation}
\mathbf{R} = UV^T
\end{equation}

Now, let the singular value decomposition of $r'$ be
$\tilde{U}\tilde{D}\tilde{V}^T$, since $\mathbf{R}' = \phi M$, we
have:

\begin{equation}\label{eq:U}
  \tilde{U} = U
\end{equation}
\begin{equation}\label{eq:V}
  \tilde{V} = V
\end{equation}
\begin{equation}\label{eq:D}
  \tilde{D} = \phi D
\end{equation}

Combine equations (\ref{eq:app_rotation}), (\ref{eq:U}) and
(\ref{eq:V}), the rotation matrix $r$ can be recovered by:

\begin{equation}\label{eq:recover_roation}
  \mathbf{R} = \tilde{U} \tilde{V}^T
\end{equation}

When noise $N$ is not significant, $D \approx I_{3\times 3}$, the
scale factor $\phi$ can be estimated by the following approximation:
\begin{equation}
trace(\tilde{D}) = trace(\phi D)\approx trace(\phi I_{3\times
3})\approx 3 \phi
\end{equation}
\begin{equation}\label{eq:phi}
\phi \approx \frac{1}{3} trace(\tilde{D})
\end{equation}

In short, if we have enough samples of $\mathbf{R}_{A}^i$,
$\hat{T}_{A}^i$, $\mathbf{R}_{B}^i$ and $\hat{T}_{B}^i$ we can find
$\hat{O}_A$, $\hat{\mathbf{R}}_{BA}$ and $\hat{T}_{BA}$(see section
\ref{sec:model_scale}). Then using the above formulas, in
particular, equation (\ref{eq:recover_roation}) and (\ref{eq:phi}),
we can also find the real rotation ($\mathbf{R}_{BA}$) and the
relative scale
factor $\phi_{BA}$.\\

Let $\mathbf{R}_{BA}=M(r,p,y)$, where $r$, $p$ and $y$ are the
corresponding roll, pitch and yaw of $\mathbf{R}_{BA}$, $M$ is a
function from roll, pitch and yaw to the corresponding rotation
matrix. In our simulation and real experiment, the estimated $r$,
$p$, $y$, $\hat{T}_{BA}$ and $\phi_{BA}$ can be optimized by
minimizing the nonlinear error function:
\begin{eqnarray}\label{eq:opt_acs_scale}
  E(r,p,y,T_{BA},Q_A) =\sum_{i=1}^n(\phi_{BA}R_B^i M(r,p,y)\hat{\bar {O}}_A + R_B^i \hat{T}_{BA}
  - \phi_{BA}M(r,p,y) R_A^i \hat{\bar{O}}_A
-\phi_{BA}M(r,p,y)T_A^i+ \hat{T}_B^i - \hat{T}_{BA})&&
\end{eqnarray}
using a Levenberg-Marquardt method.  If the pose of the joint is
calibrated with known scale factor ($O_{A}$ is known), the scale
factor $\mu_A$ can be estimated by equation (\ref{eq:scale_O_mu}).
The scale factor $\mu_B$ can be calculated by
$\frac{\mu_A}{\phi_{BA}}$. Finally, the $\mathbf{R}_{BA}$ is
recovered from the optimized $r$, $p$ and $y$. The relative pose
between the $C_{A}^{init}$ and $C_{B}^{init}$ is calculated by
equations (\ref{eq:R_define}) and (\ref{eq:T_define}). Therefore, a
non-overlapping view ACS can also be calibrated using scaled motion
information from each camera in it.
\section{Simulation} \label{sec:simulation}
In this section, the proposed calibration methods are evaluated with
synthetic transformation data.
\subsection{Performance w.r.t. Noise in Transformation Data}
\textbf{Setup and Notations:} In each test, one ACS with 2 cameras
and 1 joint is generated randomly. In which, $1\leq |O_A|\leq 2$
meters, $1\leq |O_B|\leq 2$ meters. The generated ACS performs $30$
random transformations.

\begin{figure}[tbh]
\centering
\begin{tabular}{c c}
  \includegraphics[width=2.7in]{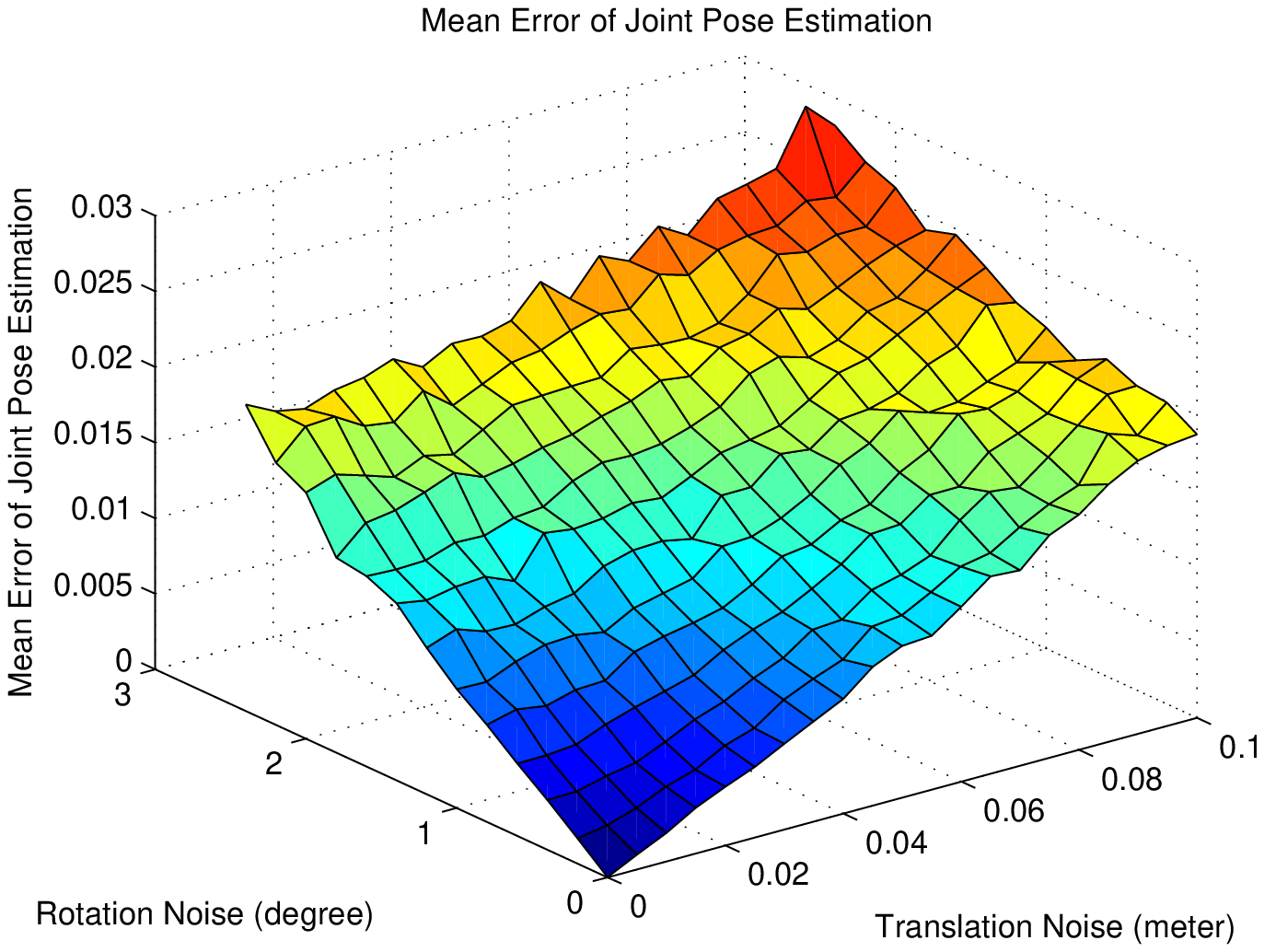} & \includegraphics[width=2.7in]{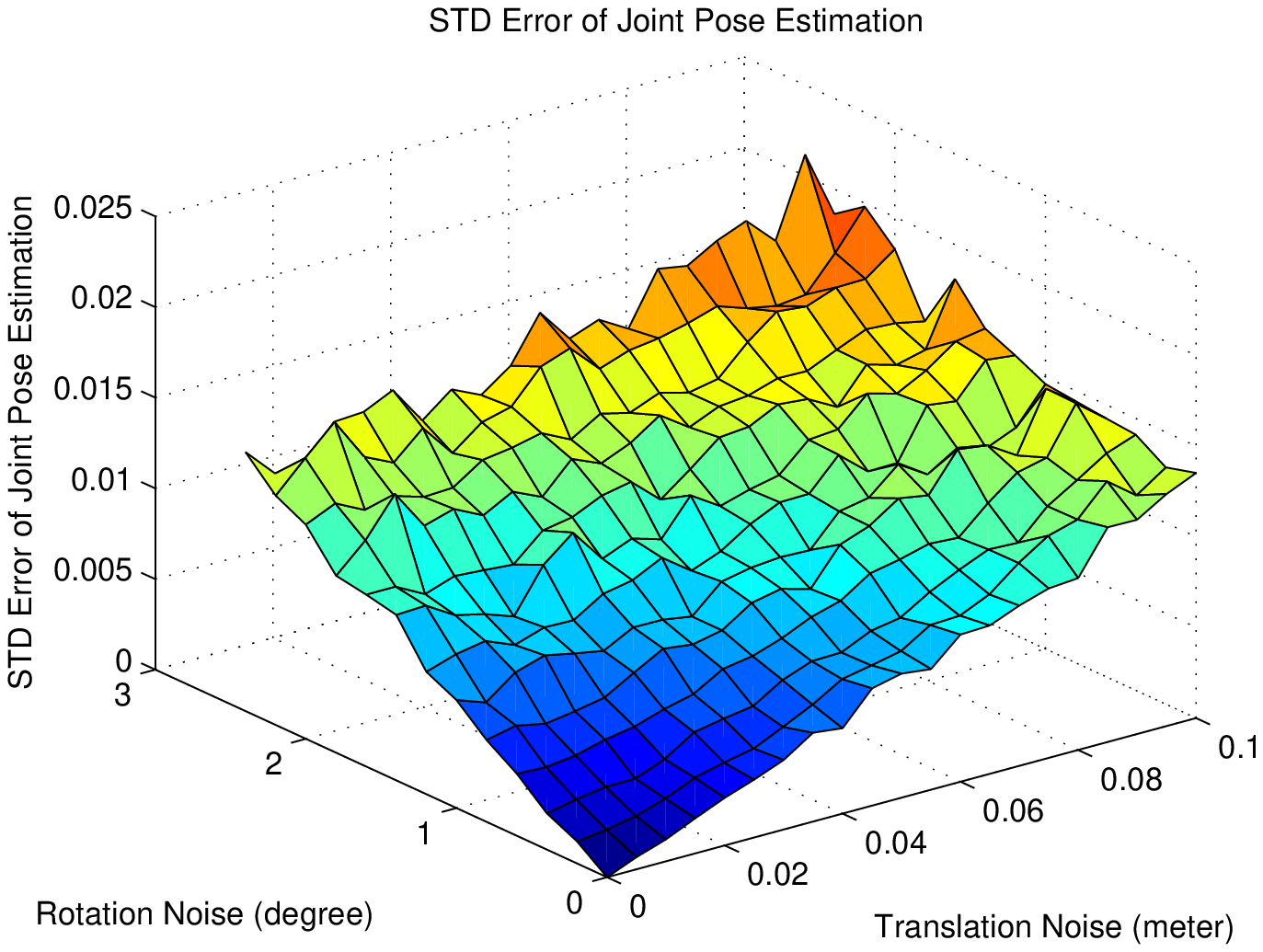}\\
(a) & (b)
\end{tabular}
\caption{Mean and STD Error of  Joint Pose ($O_A$) Estimation. (a)
Mean Error of Joint Pose Estimation; (b) STD Error of Joint
Pose.}\label{fig:joint} \end{figure}

\textbf{Performance of the Calibration Method for ACS with
Overlapping Views:} In the first simulation, the proposed algorithm
is tested 100 times. Zero mean Gaussian noise is added to the
transformation data of the cameras. The configuration, input and
output of our simulation system are list as Table \ref{tab:cio1}.
Since we assume there are overlapping views between the two cameras,
the relative pose between them can be estimated by many existing
methods as discussed in section \ref{sec:introduction}. Only the
performance of joint pose estimation is evaluated in our simulation.
The error of joint estimation are computed by:

\begin{equation}\label{eq:err}
  Err = \frac{|\bar{O}_A - \hat{\bar{O}}_A|}{2|\bar{O}_A|} + \frac{|\bar{O}_B - \hat{\bar{O}}_B|}{2|\bar{O}_B|}
\end{equation}
, where $\bar{O}_A$ is the ground truth, $\hat{\bar{O}}_A$ is the
estimated position of joint O relative to camera A. Similarly,
$\bar{O}_B$ is the ground truth, $\hat{\bar{O}}_B$ is the estimated
position of joint O relative to camera B.
The corresponding results are shown in Figure \ref{fig:joint}.
\begin{table}[h]
    \centering
    \caption{Configuration, Input and Output}\label{tab:cio1}
    \begin{tabular}{ll}
    \hline
    \multicolumn{2}{c}{\textbf{Configuration}}\\
    \hline
    No. of Cameras in the ACS & 2\\
    No. of Joints in the ACS & 1\\
    {Random transformations per test} (n) & 30\\
    Number of tests & 100\\
    \hline
    \multicolumn{2}{c}{\textbf{Input ($i = 1 \dots n$)}}\\
    \hline
    Rotations of cameras ($\mathbf{R}_{AW}^i$, $\mathbf{R}_{BW}^i$) & $2\times 30\times 100$\\
    Translations of cameras ($T_{AW}^i$, $T_{BW}^i$) & $2\times 30\times 100$\\
    \multicolumn{2}{l}{\textbf{Zero Mean Gaussian noise:}}\\
    \multicolumn{2}{l}{$0\leq\sigma_{rot} \leq 2.4^{\circ}$  and $0\leq\sigma_{trans} \leq 0.1 meters$}\\

    \hline
    \multicolumn{2}{c}{\textbf{Output}}\\
    \hline
    \multicolumn{2}{l}{Mean error of joint pose estimation (see equation (\ref{eq:err}))}\\
    \multicolumn{2}{l}{STD error of joint pose estimation (see equation (\ref{eq:err}))}\\
    \hline
    \end{tabular}
\end{table}

\textbf{Performance of the Calibration Method for Non-Overlapping
Views ACS:} In the second simulation, firstly, the pose of the joint
is fixed relative to $C_A^{init}$ during the transformations of the
ACS. The pose of the joint relative to the camera A ($O_A$) is
calibrated by the transformations of camera A. Similarly, $O_B$ is
calibrated. Then, the ACS performs several general transformations
(the joint is not needed to be fixed relative to $C_A^{init}$), the
relative pose between the cameras are calibrated using the estimated
joint pose and the transformations of the cameras. The
configuration, input and output of the simulation system are listed
as Table \ref{tab:cio2}. The error of joint pose, relative rotation,
relative translation estimation are calculated by equation
(\ref{eq:err}), (\ref{eq:rot_err}) and (\ref{eq:trans_err})
respectively.

\begin{figure}[tbh]
\centering
\begin{tabular}{c c}
  \includegraphics[width=2.78in]{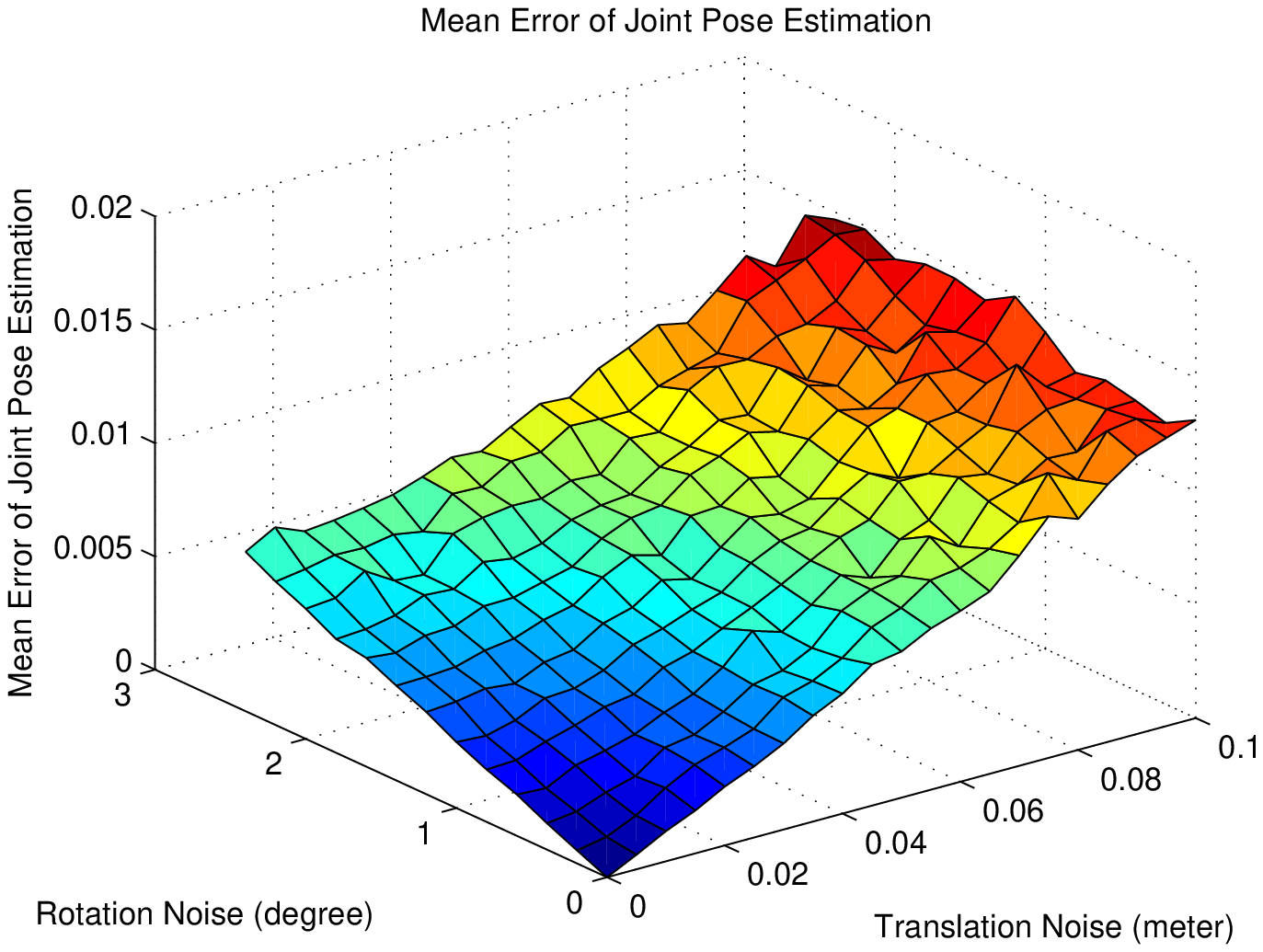} & \includegraphics[width=2.78in]{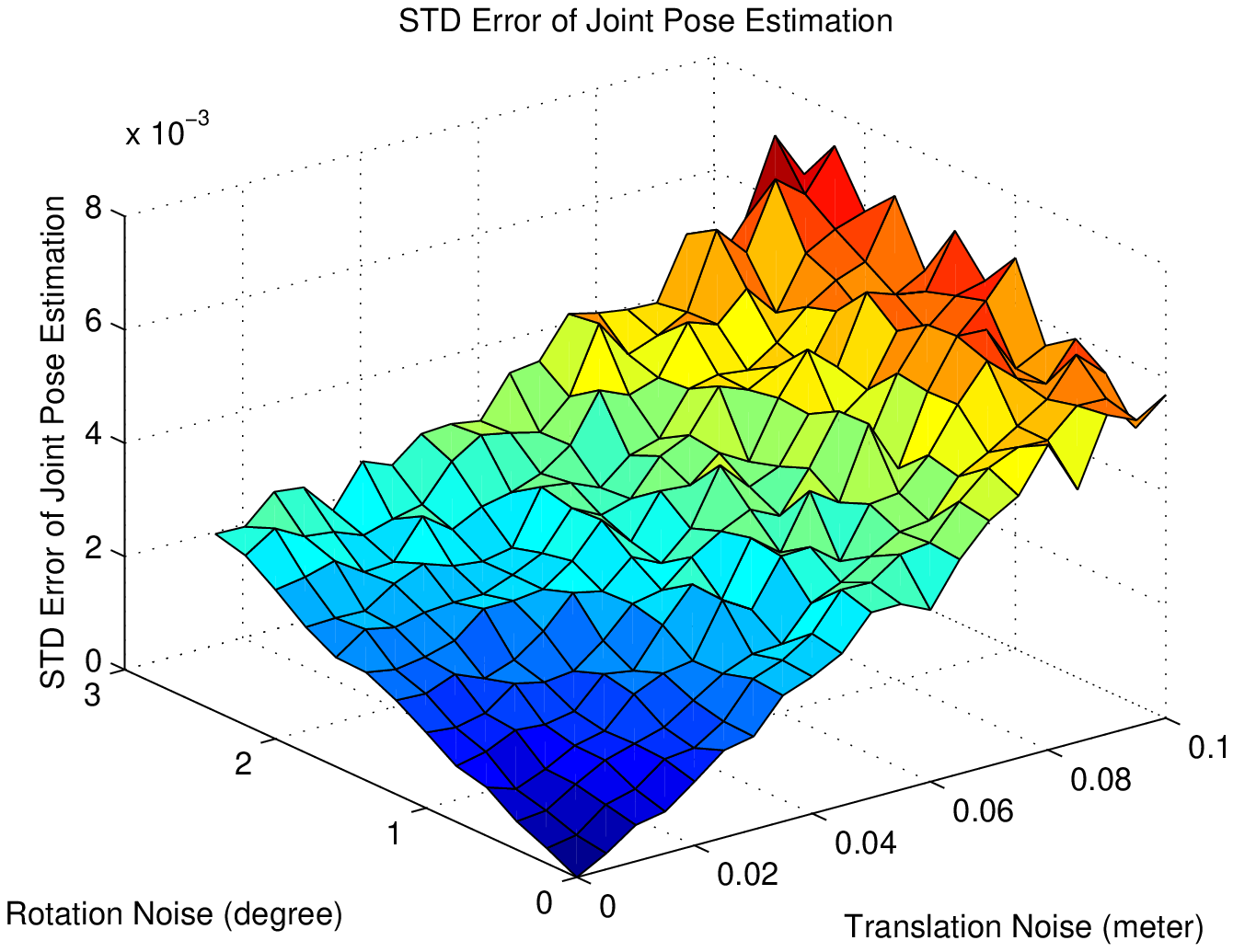}\\
(a) & (b)
\end{tabular}
\caption{Mean and STD Error of  Joint Pose ($\hat{O}_A$) Estimation.
(a) Mean Error of Joint Pose Estimation; (b) STD Error of Joint Pose
Estimation.}\label{fig:joint_non_overlap}
\end{figure}

\begin{figure}[tbh]
\centering
\begin{tabular}{c c}
  \includegraphics[width=2.78in]{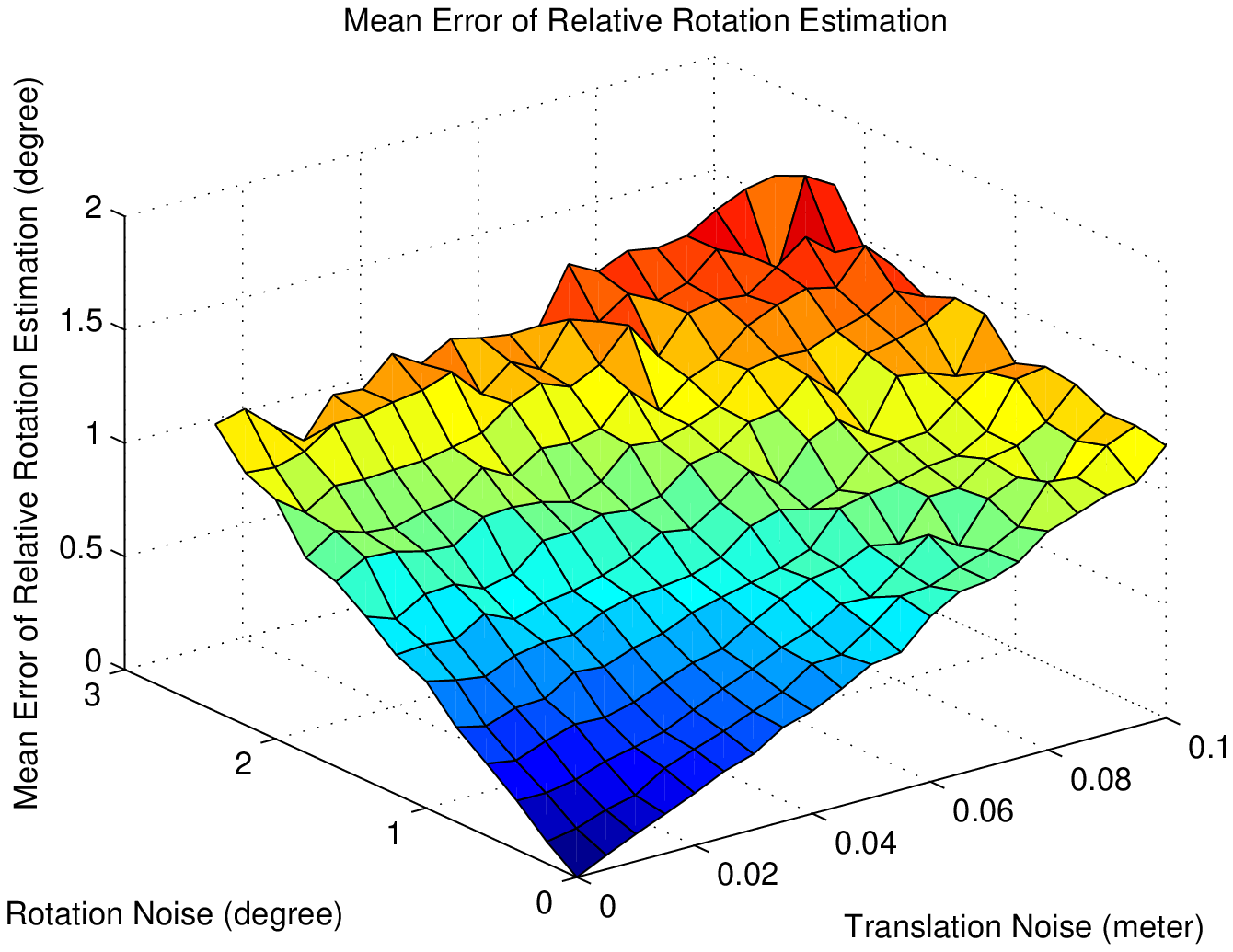} & \includegraphics[width=2.78in]{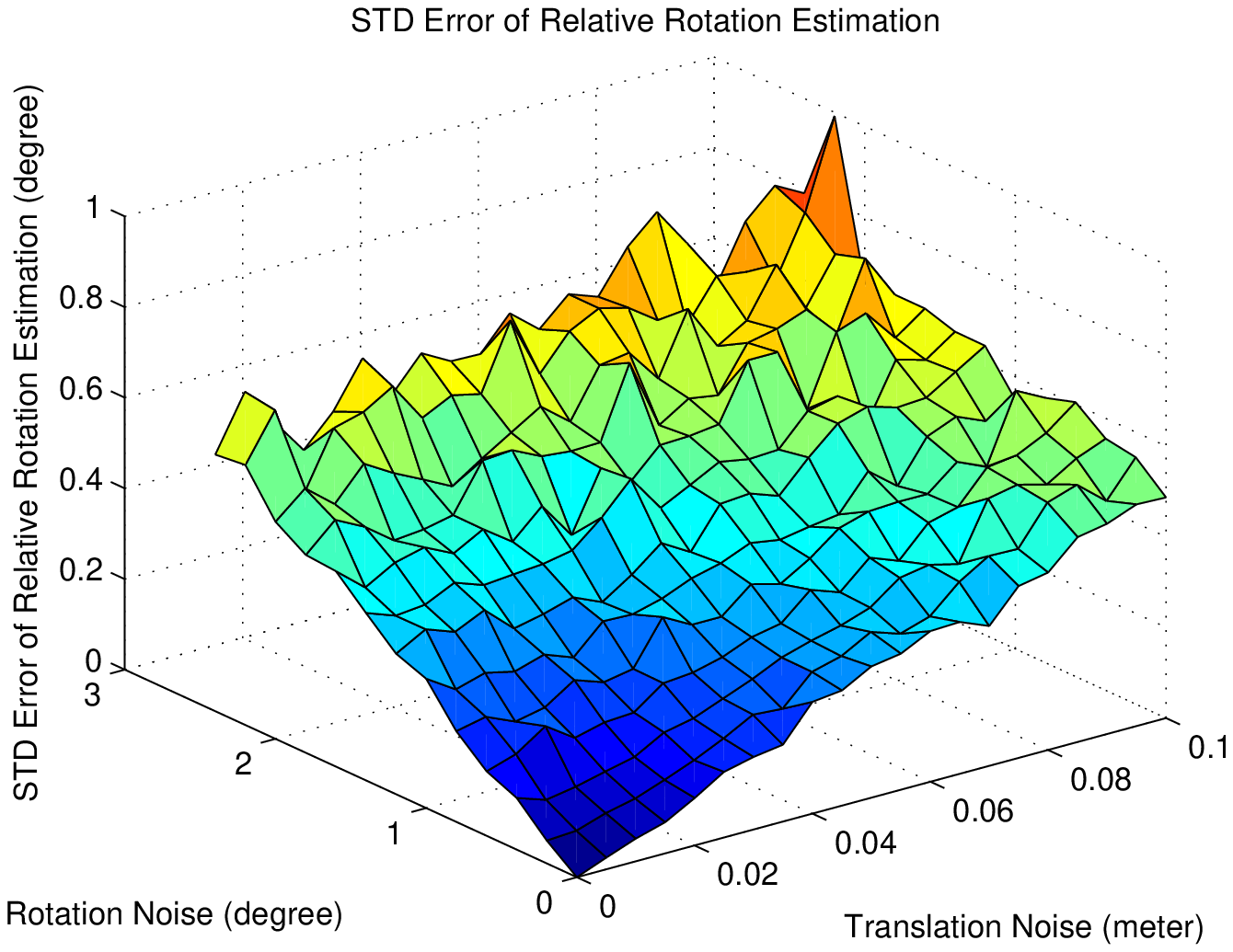}\\
(a) & (b)
\end{tabular}
\caption{Mean and STD Error of  Relative Rotation  ($R_{BA}$)
Estimation. (a) Mean Error of Relative Rotation Estimation; (b) STD
Error of Relative Rotation Estimation.}\label{fig:rot_non_overlap}
\end{figure}

\begin{figure}[tbh]
\centering
\begin{tabular}{c c}
  \includegraphics[width=2.78in]{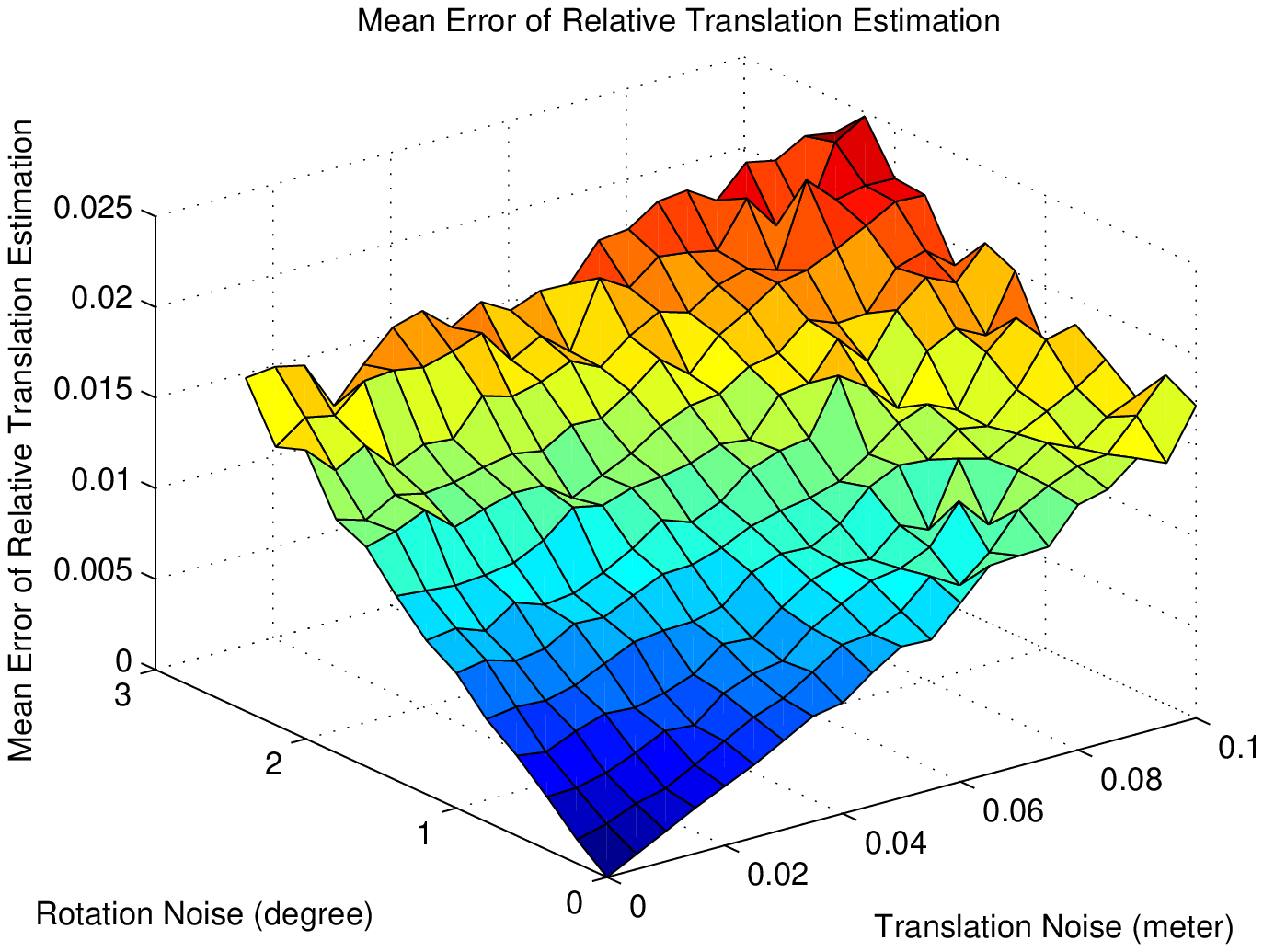}&  \includegraphics[width=2.78in]{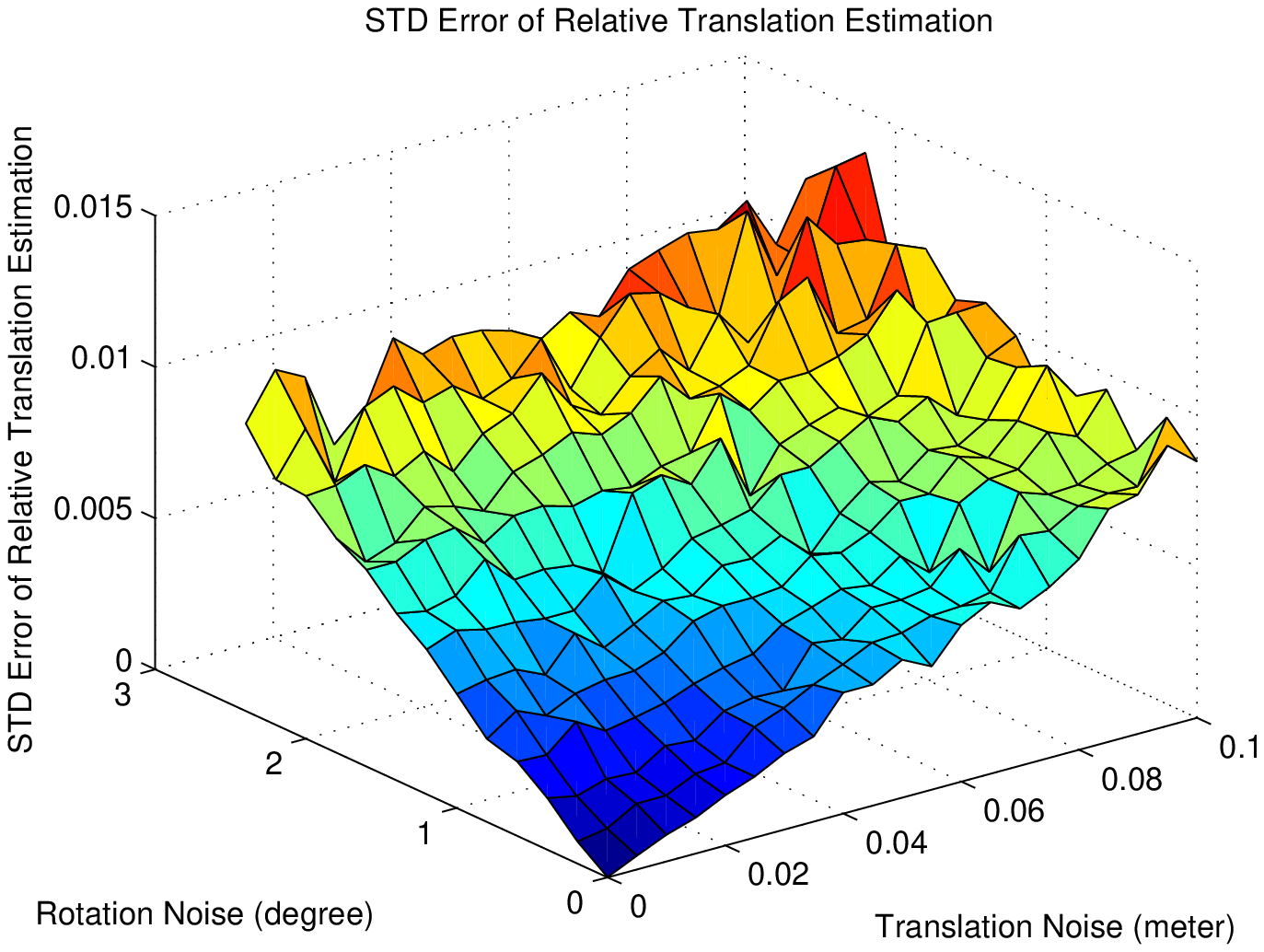}\\
(a) & (b)
\end{tabular}
\caption{Mean and STD Error of Relative Translation ($\hat{T}_{BA}$)
Estimation. (a) Mean Error of Relative Translation Estimation; (b)
STD Error of Relative Translation Estimation.}
\label{fig:trans_non_overlap}
\end{figure}

Figure \ref{fig:joint_non_overlap} shows the results of joint pose
estimation. Compare with the calibration method using the
overlapping views, the calibration method using special motions is
more accurate. The mean and STD error of the relative rotation and
translation estimation are presented in Figure
\ref{fig:rot_non_overlap} and \ref{fig:trans_non_overlap}. The
proposed algorithms are shown to be stable, when the zero mean
Gaussian noise from $0^\circ$ to $2.4^\circ$ is added to the roll,
pitch and yaw of the rotation data, and the zero mean Gaussian noise
from $0$ to $0.1$ meters is added to the translation data.

\begin{equation}\label{eq:rot_err}
Err^{rot} = \sqrt{|roll - \widehat{roll}|^2 +|pitch -
\widehat{pitch}|^2+|yaw - \widehat{yaw}|^2}
\end{equation}
\begin{equation}\label{eq:trans_err}
Err^{trans} = \frac{|T_{AB} - \hat{T}_{AB}|}{|T_{AB}|}
\end{equation}
\begin{table}[htb]
    \centering
    \caption{Configuration, Input and Output}\label{tab:cio2}
    \begin{tabular}{ll}
    \hline
    \multicolumn{2}{c}{\textbf{Configuration}}\\
    \hline
    No. of Cameras in the ACS & 2\\
    No. of Joints in the ACS & 1\\
    {Random transformations per test} (n) & 30\\
    Number of tests & 100\\
    \hline
    \multicolumn{2}{c}{\textbf{Input ($i = 1 \dots n$)}}\\
    \hline
    \multicolumn{2}{l}{\textbf{Transformations with fixed joint pose:}}\\
    Rotations of cameras ($\mathbf{R}_{A}^i$, $\mathbf{R}_{B}^i$) & $2\times 30\times 100$\\
    Translations of cameras ($T_{A}^i$, $T_{B}^i$) & $2\times 30\times 100$\\
    \multicolumn{2}{l}{\textbf{General transformations:}}\\
    Rotations of cameras ($\mathbf{R}_{A}^i$, $\mathbf{R}_{B}^i$) & $2\times 30\times 100$\\
    Translations of cameras ($T_{A}^i$, $T_{B}^i$) & $2\times 30\times 100$\\
    \multicolumn{2}{l}{\textbf{Zero Mean Gaussian noise:}}\\
    \multicolumn{2}{l}{$0\leq\sigma_{rot} \leq 2.4^{\circ}$  and $0\leq\sigma_{trans} \leq 0.1 meters$}\\
    \hline
    \multicolumn{2}{c}{\textbf{Output}}\\
    \hline
    \multicolumn{2}{l}{Mean error of joint pose estimation (see equation (\ref{eq:err}))}\\
    \multicolumn{2}{l}{STD error of joint pose estimation (see equation (\ref{eq:err}))}\\
    \multicolumn{2}{l}{Mean error of relative translation estimation (see equation (\ref{eq:trans_err}))}\\
    \multicolumn{2}{l}{STD error of relative translation estimation (see equation (\ref{eq:trans_err}))}\\
    \multicolumn{2}{l}{Mean error of relative rotation estimation (see equation (\ref{eq:rot_err}))}\\
    \multicolumn{2}{l}{STD error of relative rotation estimation (see equation (\ref{eq:rot_err}))}\\
    \hline
    \end{tabular}
\end{table}

\textbf{Performance of the Calibration Method for Non-Overlapping
Views ACS with Unknown Scale Factors:} The scale factors of the two
cameras in each test are assumed to be uniform distributed in the
range $[0.5,5]$. Therefore, the relative scale factor between the
two cameras satisfies the uniform distribution in the range of
$[0.1, 10]$. The joint pose of the ACS is generate randomly and
estimated by the method described in section
\ref{sec:special_motion}.
Other configurations are the same as the second simulation. The
$\hat{O}_A$, $R_{BA}$, $\hat{T}_{BA}$ and $\phi_{BA}$ are estimated
and optimized as discussed in section \ref{sec:deal_scale}. The
error of joint pose, relative rotation, relative translation
estimation are calculated by equation (\ref{eq:err}),
(\ref{eq:rot_err}) and (\ref{eq:trans_err}) respectively. The error
of relative scale factor estimation is evaluated by
$\varepsilon_{\phi} = \frac{|\phi - \hat{\phi|}}{|\phi|}$. Where
$\hat{\phi}$ is the estimated relative scale factor, and $\phi$ is
the ground truth. \\

\begin{figure}[tbh]
\centering
\begin{tabular}{c c}
  \includegraphics[width=2.78in]{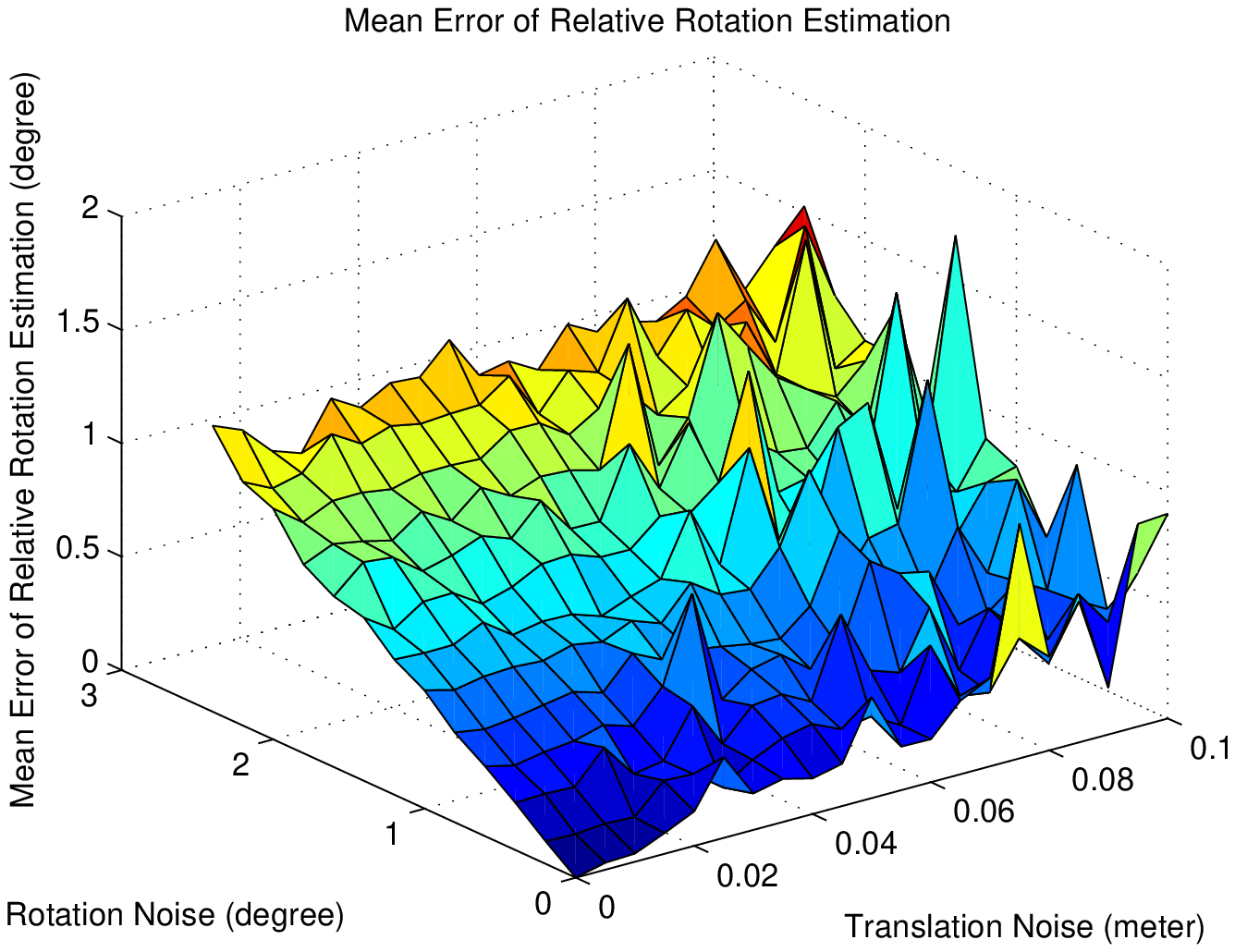} &   \includegraphics[width=2.78in]{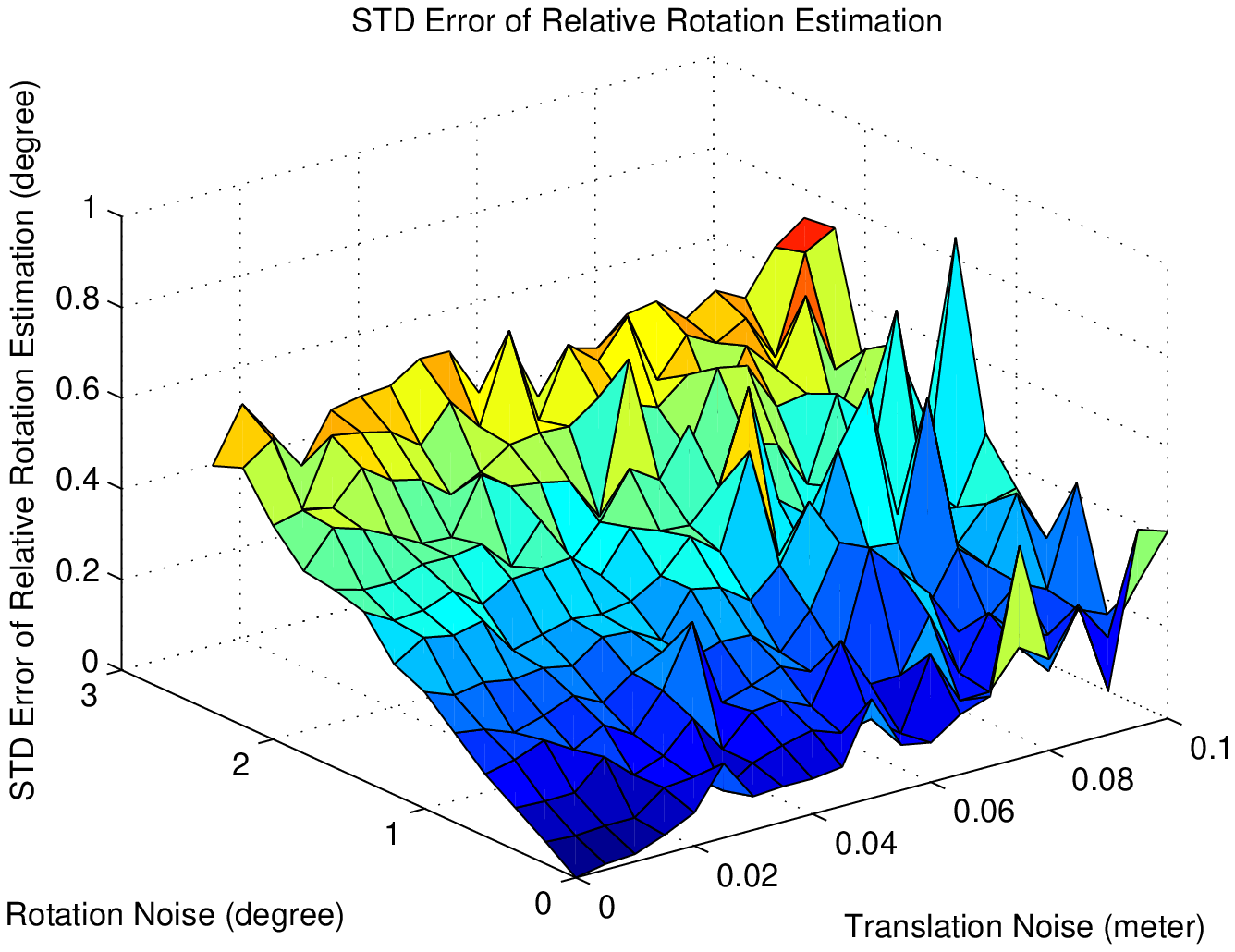}\\
(a) & (b)
\end{tabular}
\caption{Mean and STD Error of  Relative Rotation with Unknown Scale
Factor $(R_{BA})$. (a) Mean Error of Relative Rotation; (b) STD
Error of Relative Rotation.}\label{fig:rot_scale}
\end{figure}

\begin{figure}[tbh]
\centering
\begin{tabular}{c c}
  \includegraphics[width=2.78in]{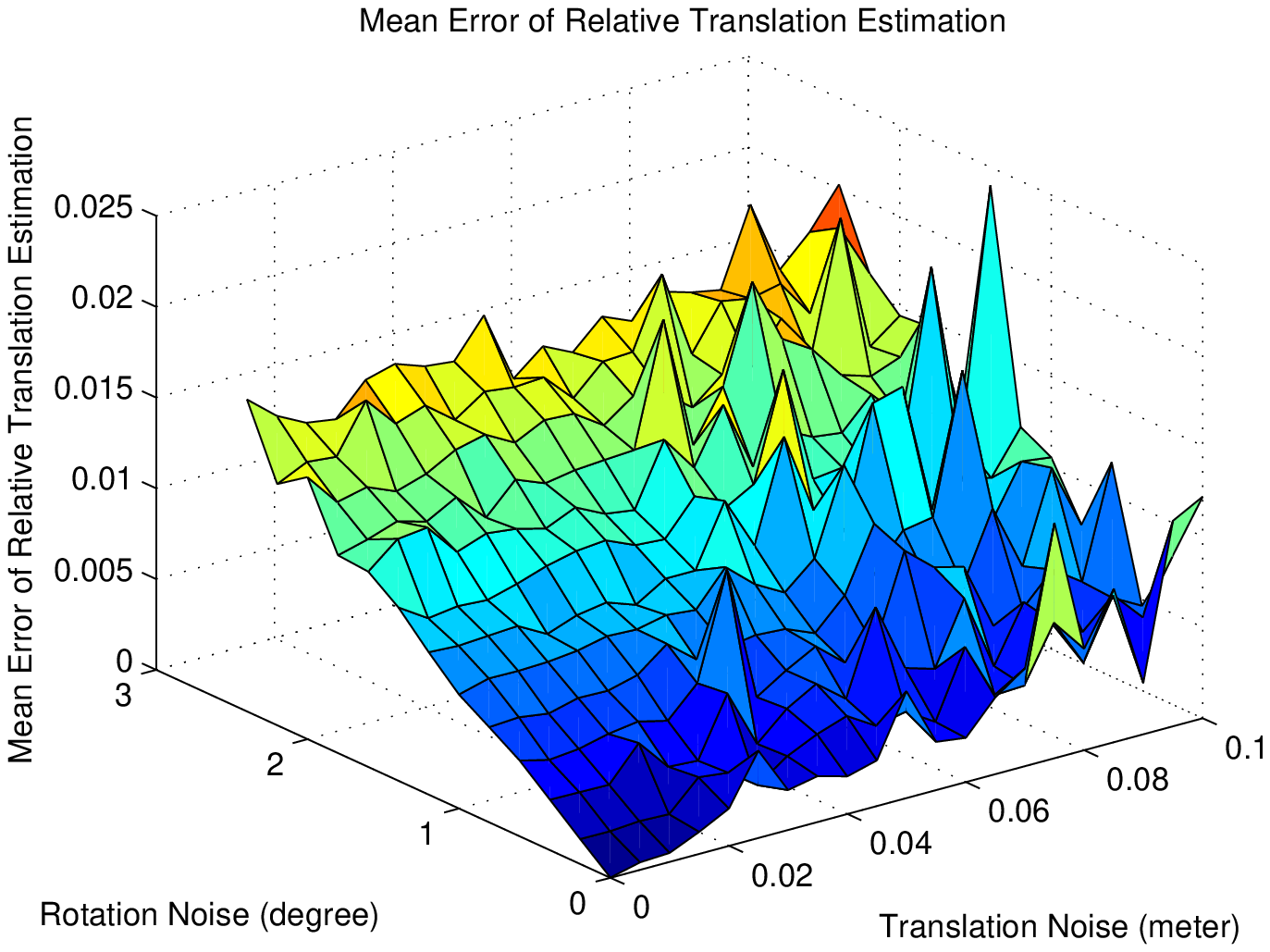}&   \includegraphics[width=2.78in]{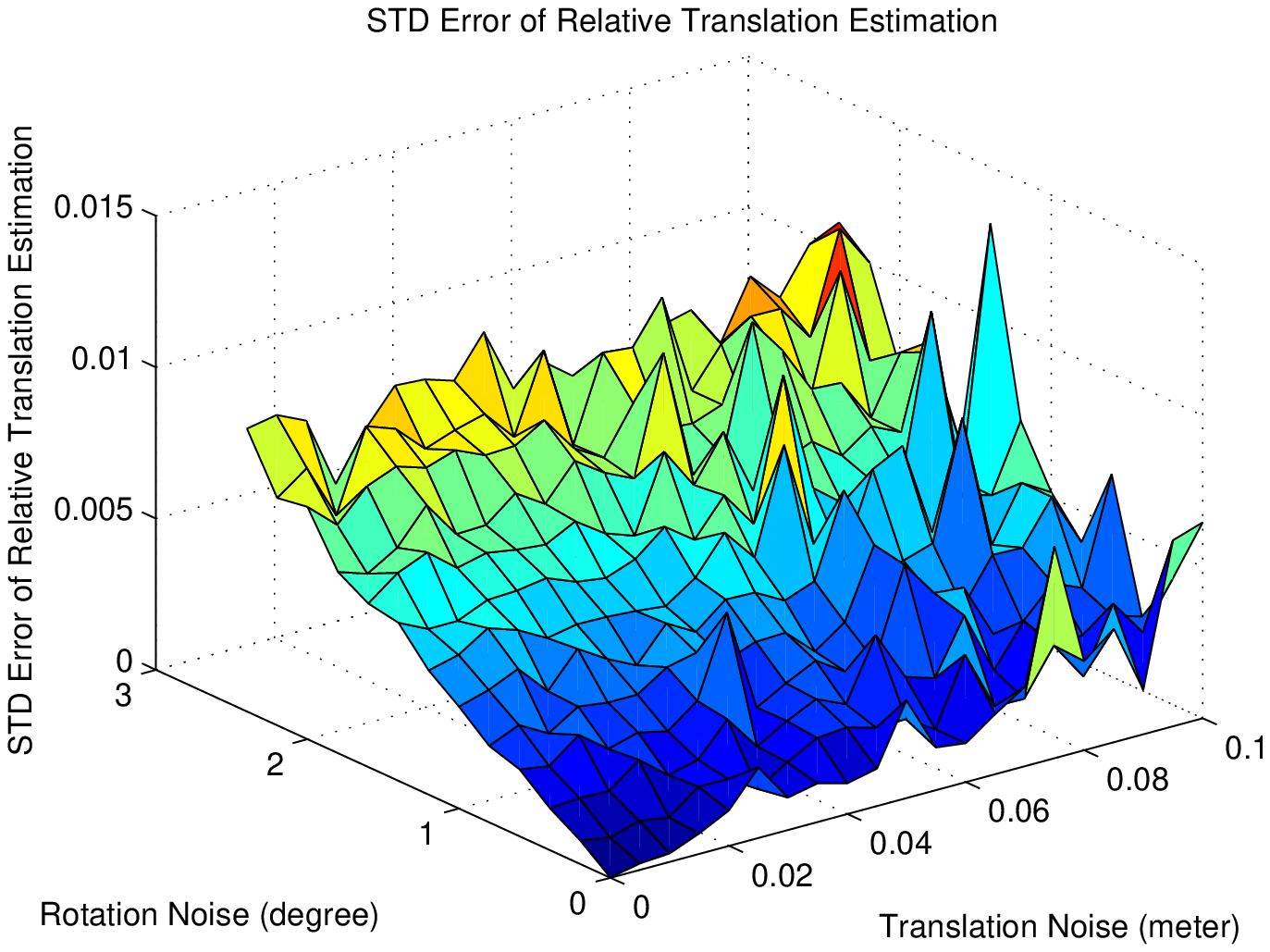}\\
(a) & (b)
\end{tabular}
\caption{Mean and STD Error of Relative Translation with Unknown
Scale Factor ($\hat{T}_{BA}$). (a) Mean Error of Relative
Translation; (b) STD Error of Relative Translation.}
\label{fig:trans_scale}
\end{figure}

Figure \ref{fig:rot_scale} and \ref{fig:trans_scale} show the
results of the relative pose estimation. Compared to figure
\ref{fig:rot_non_overlap} and \ref{fig:trans_non_overlap} the
accuracies are similar.

\begin{figure}[tbh]
\centering
\begin{tabular}{c c}
  \includegraphics[width=2.78in]{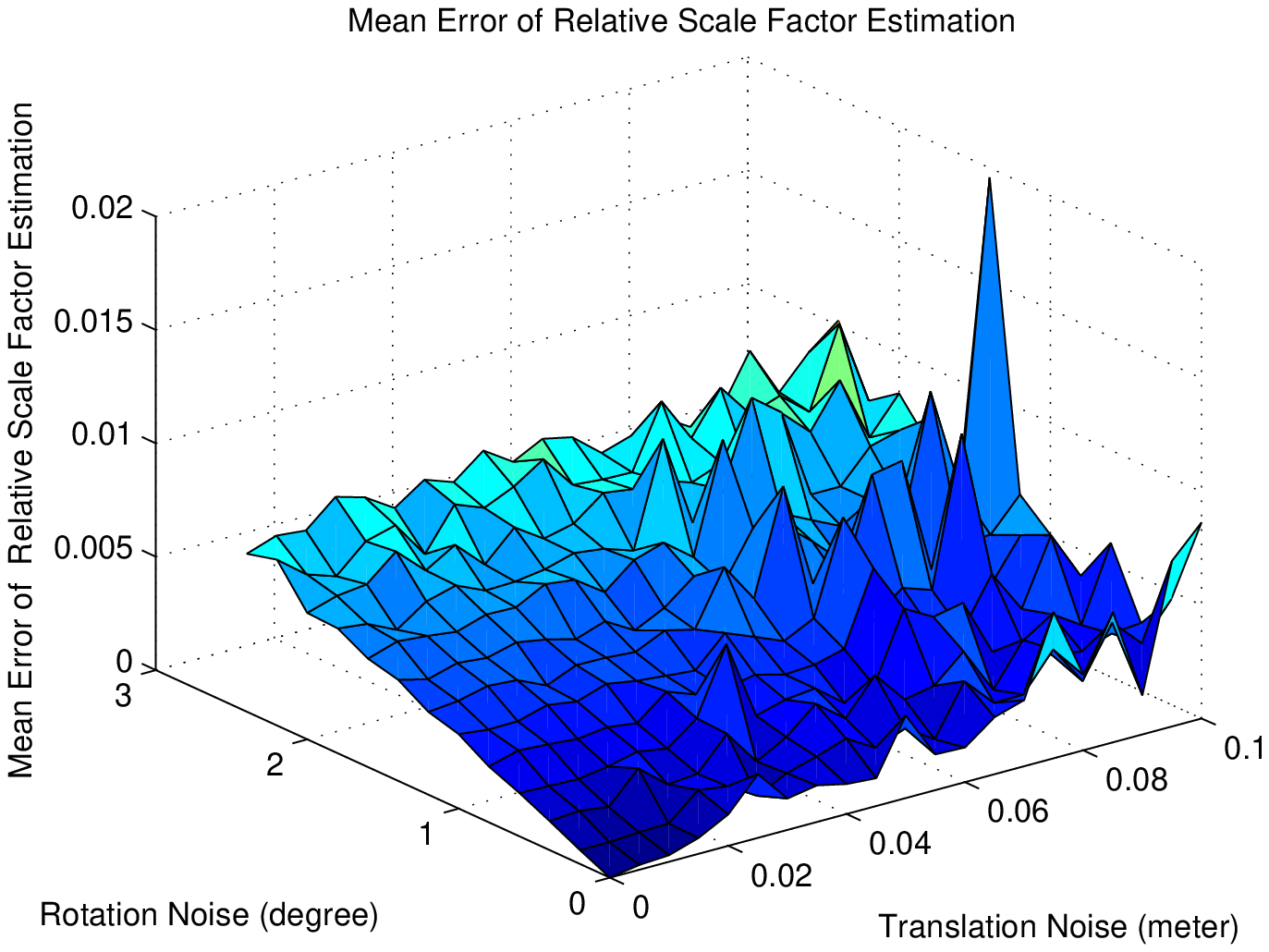}&   \includegraphics[width=2.78in]{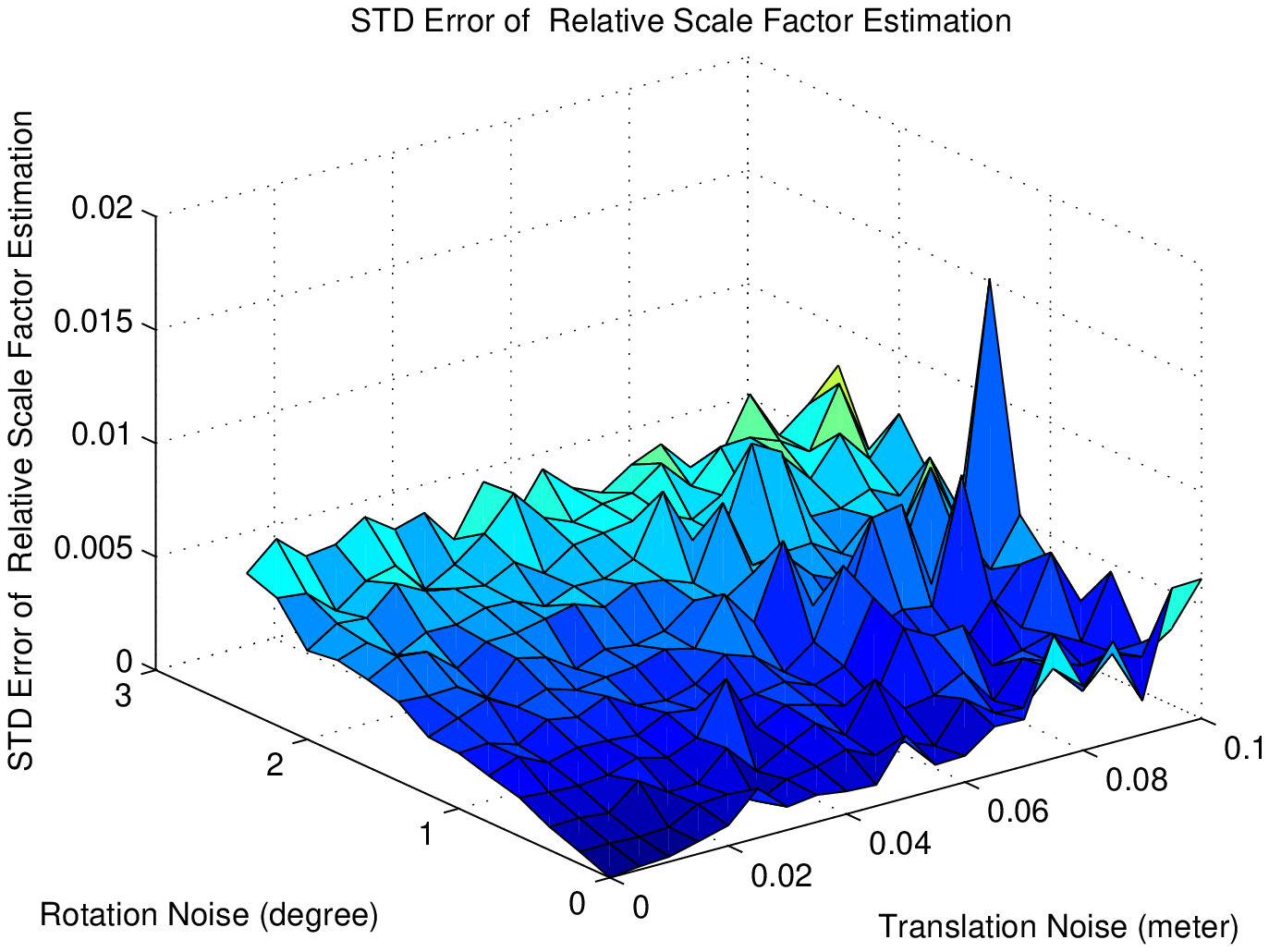}\\
(a) & (b)
\end{tabular}
\caption{Mean and STD Error of  Relative Scale Factor with Unknown
Scale Factor ($\phi_{BA}$). (a) Mean Error of Relative Scale Factor;
(b) STD Error of Relative Scale Factor.}\label{fig:scale_factor}
\end{figure}

Figure \ref{fig:scale_factor} shows the performance of the relative
scale factor estimation. The accuracy of the relative scale factor
estimation $[(1-\varepsilon_\phi)\times 100\%]$ is no less than
$98.5\%$, when the standard derivation of the noise in ego-rotation
is less than $3^{\circ}$ and the standard derivation of the noise in
ego-translation is less than $0.1$ meters.\\

\begin{figure}[tbh]
\centering
\begin{tabular}{c c}
  \includegraphics[width=2.78in]{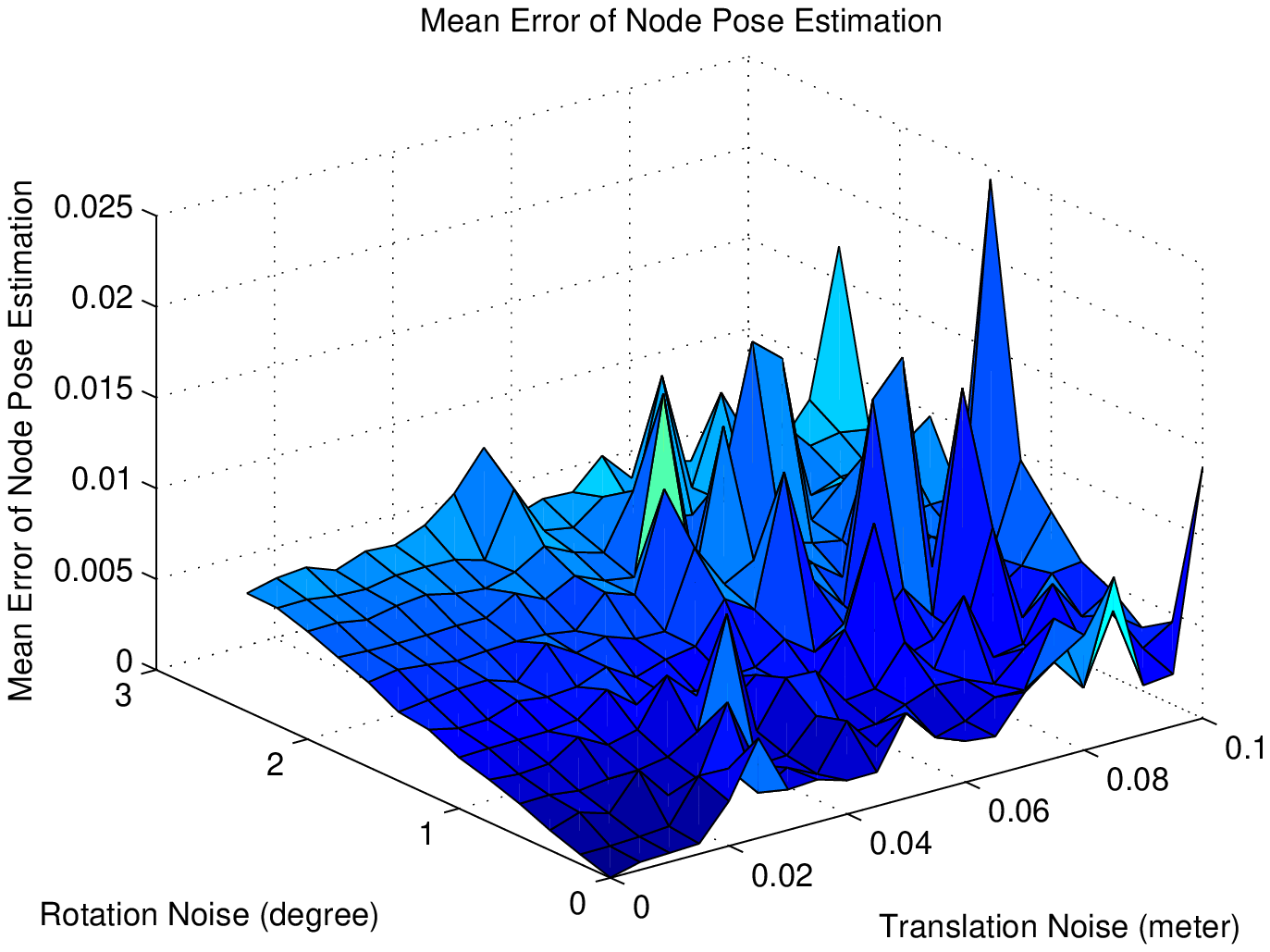} &   \includegraphics[width=2.78in]{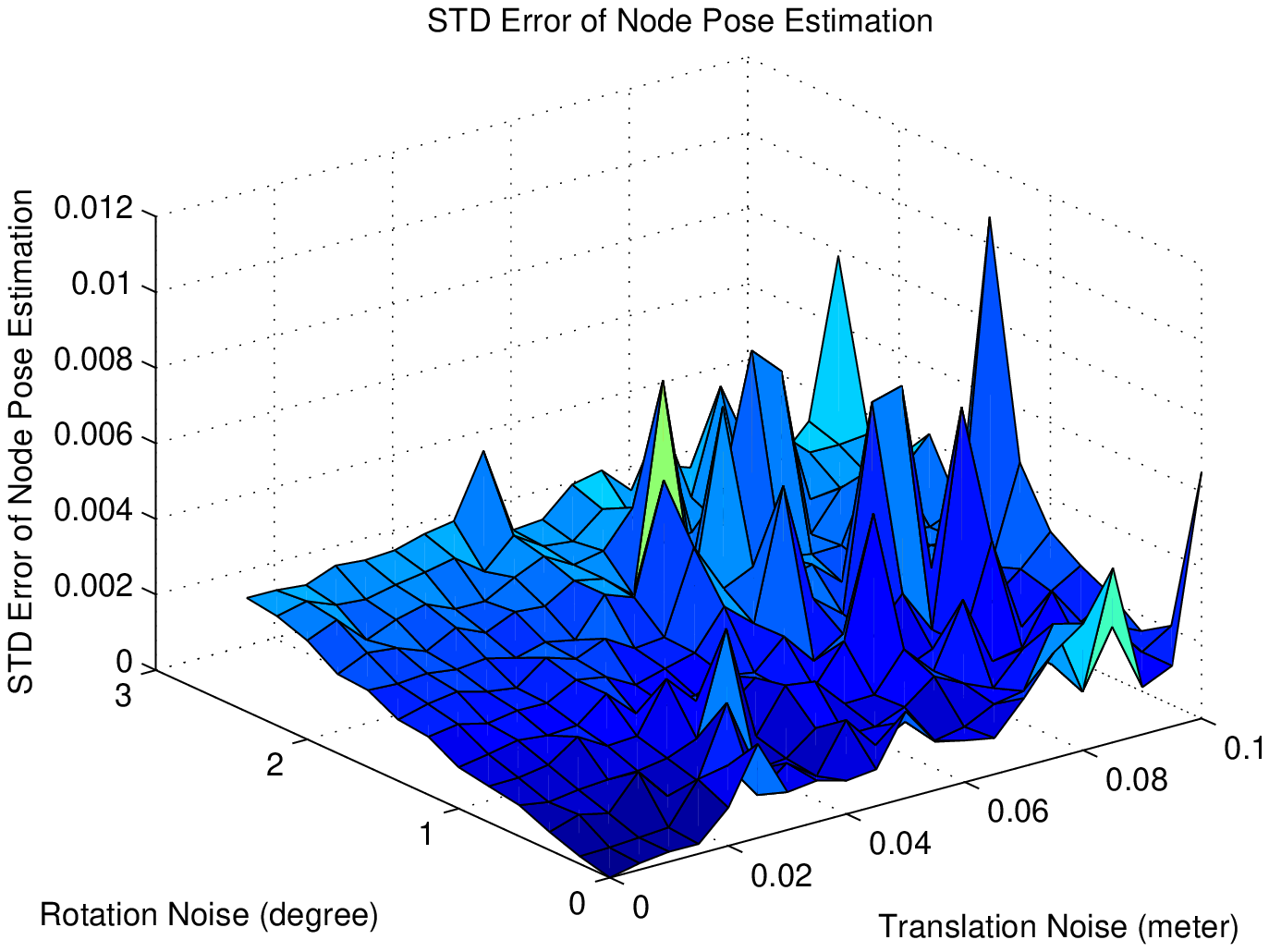}\\
(a) & (b)
\end{tabular}
\caption{Mean and STD Error of  Joint Pose with Unknown Scale Factor
($\hat{O}_A$). (a) Mean Error of Joint Pose; (b) STD Error of Joint
Pose.}\label{fig:joint_scale} \end{figure}

\section{Real Experiment} \label{sec:real}
In the real experiments, an ACS with two cameras (\emph{Cannon
PowerShot G9}) is set up as Figure \ref{fig:non_overlap_cfg} (a).
The intrinsic parameters of each camera in the ACS are calibrated by
Bouguet's implementation (``Camera Calibration Toolbox for Matlab'')
of \cite{zhang00flexible}. Since the Bouguet's Toolbox can also
estimate the pose information of the camera, the transformations of
each camera are calculated using the same image sequence for the
intrinsic calibration simultaneously. No additional images nor
manual input is required in the real experiments.

\subsection{Calibration of the Pose of the Joint in Each
Camera}\label{sec:alg_i_ii} \textbf{By Overlapping Views (Algorithm
I):} In the first real experiment, the two cameras in the ACS
observe the same checker plane and record images simultaneously. The
two cameras are free to move during the transformation of the ACS.
Two image sequences ($Q_1$ and $Q_2$) are recorded, each sequence
consists of $15$ images of size $1600\times 1200$ pixels. The
estimated joint pose are list in Table \ref{tab:joint_pose} as
algorithm I.

\textbf{By Fixed-Joint Motions (Algorithm II):} In the second real
experiment, the joint of the ACS is fixed relative to the world
coordinate system during the transformation of the ACS. The two
cameras do not need to view the same checker plane. And each camera
records the image sequence independently. Two image sequences ($Q_3$
and $Q_4$) are recorded, each sequence consists of $12$ images of
size $1600\times 1200$ pixels. The camera pose of the first image is
selected as the initial pose to generate the transformation sequence
of each camera. The estimated joint pose are list in Table
\ref{tab:joint_pose} as algorithm II. The poses of the joint
relative to the two cameras in the ACS are also estimated manually
for comparison purpose. Since the camera pose of any image in each
image sequence can be chosen as the initial camera pose (see section
\ref{sec:special_motion}), the proposed algorithm is also tested by
choosing different images as the reference. The mean and standard
derivation of the corresponding calibration results are presented in
Table \ref{tab:joint_ms}.
\begin{table}[h]
\centering  \caption{Results Of Joint Pose
Calibration}\label{tab:joint_pose}
{\flushleft \footnotesize I: the algorithm using overlapping views.
(see section \ref{sec:alg_i_ii}) II: the algorithm using fixed-joint
motions. (see section \ref{sec:alg_i_ii}) M: manual
measurement(ground truth). $O_A$ is the coordinate of the joint
relative to camera A, the same applies to $O_B$.\\}

\ \\ \

{
\begin{tabular}{|c|c|c|c|c|}
\hline \multicolumn{2}{|l|}{Algorithm}
&\multicolumn{3}{c|}{Joint Pose (mm)}\\
\cline{3-5}
\multicolumn{2}{|c|}{}& X& Y&Z\\
\hline I & ${O}_A$& 300.28 &  50.07 & -33.47\\
\cline{2-5} & ${O}_B$ & -273.70 &  53.81 & -30.15\\
\hline II & ${O}_A$ &304.55 &  47.64 & -37.66\\
\cline{2-5} & ${O}_B$& -265 & 54.41 &  -35.48\\
\hline M & ${O}_A$ &300 $\pm$ 10 &  50$\pm$ 10& -40$\pm$ 10\\
\cline{2-5} & ${O}_B$& -270$\pm$ 10&  50$\pm$ 10& -30$\pm$ 10\\
\hline
\end{tabular}}
\end{table}
\begin{table}[h]
\centering  \caption{Mean and STD of the Joint Pose Calibration
Algorithm II Using Different Reference Images}\label{tab:joint_ms}
{$O_A$ is the coordinate of the joint relative to camera A, the same
applies to $O_B$.\\}
\begin{tabular}{|c|c|c|c|c|}
\hline \multicolumn{2}{|l|}{Algorithm}
&\multicolumn{3}{c|}{Joint Pose (mm)}\\
\cline{3-5}
\multicolumn{2}{|c|}{II}& X& Y&Z\\
\hline Mean & ${O}_A$& 305.44 & 47.19&-39.2\\
\cline{2-5} & ${O}_B$ &  -262.97&  56.21& -39.20\\
\hline STD & ${O}_A$ & 1.89  &  1.16  &  3.02\\
\cline{2-5} & ${O}_B$&  3.3&  2.67&  2.58\\
\hline
\end{tabular}
\end{table}
\subsection{Calibration of Relative Pose Between the Cameras in the Non-Overlapping View ACS (Algorithm
III)}\label{sec:non_overlap_exp}\label{sec:alg_iii} 
\begin{figure}[t]
\centering
  \includegraphics[height=2.7in, width = 3in]{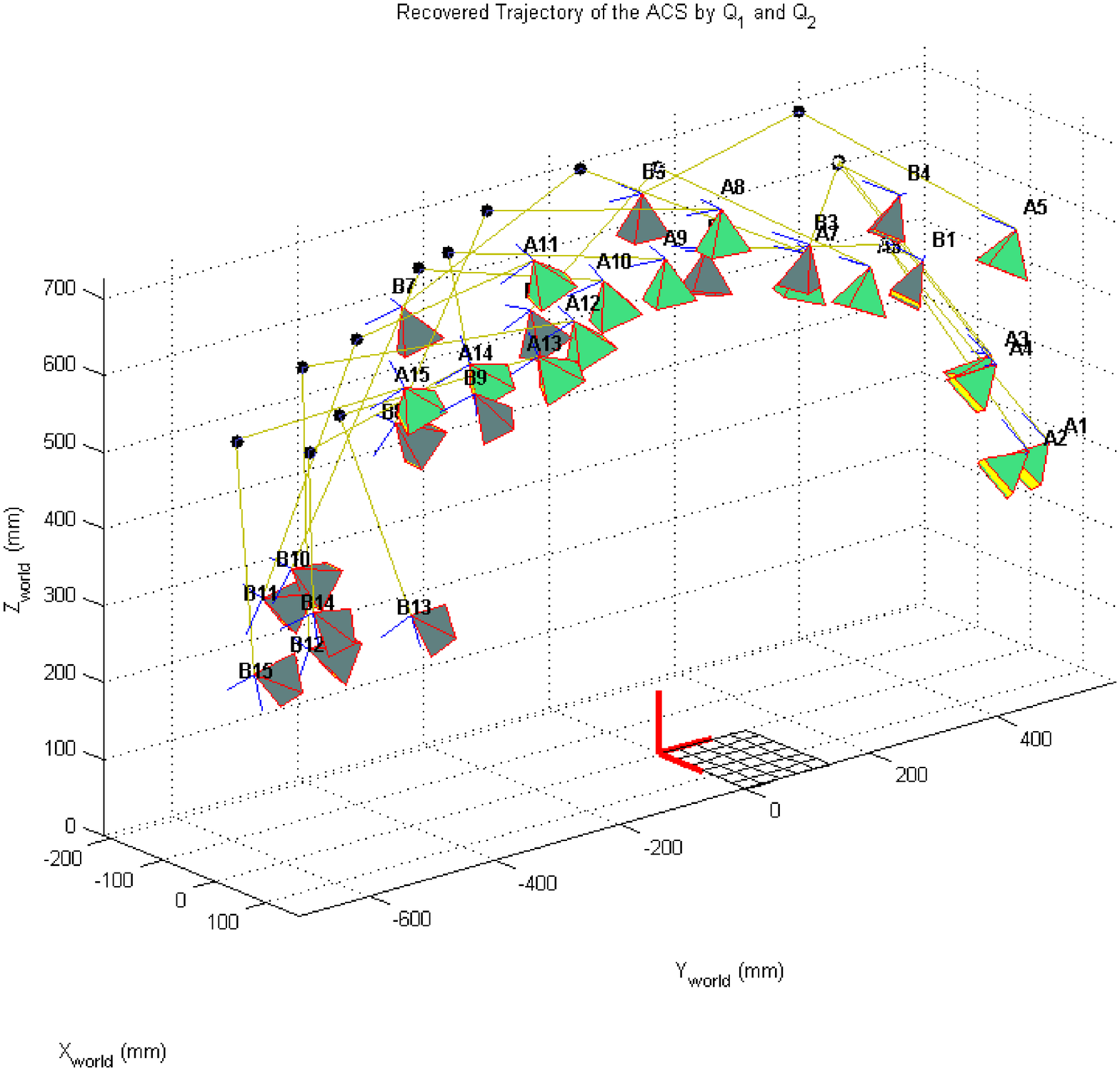}\\
\caption{The Trajectory of the ACS Recovered from $Q_1$ and $Q_2$}\label{fig:traj_overlap}
\end{figure}

\begin{figure}
\centering
\begin{tabular}{c c}
  \includegraphics[width=1.1in]{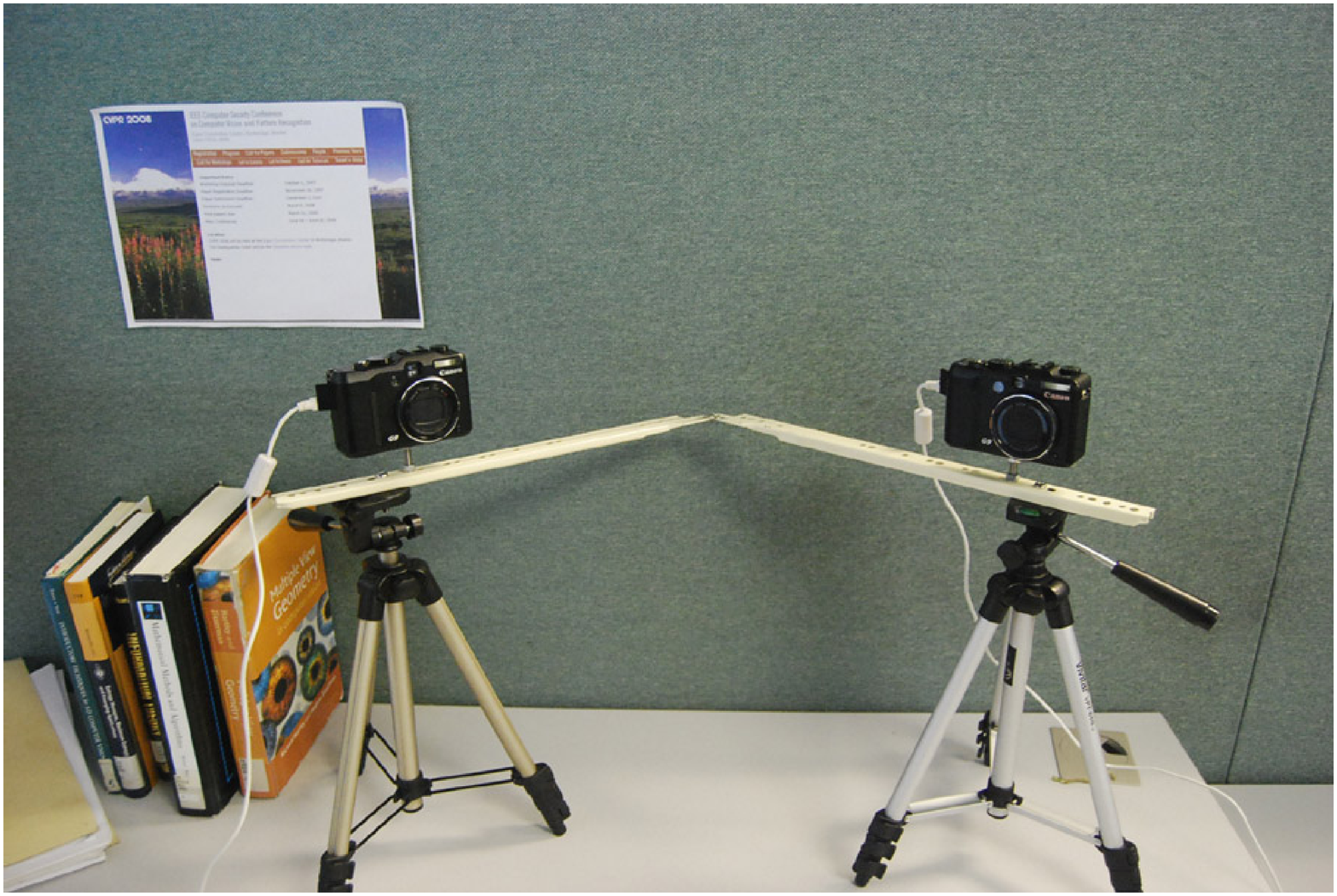}&
{\includegraphics[width=1.1in]{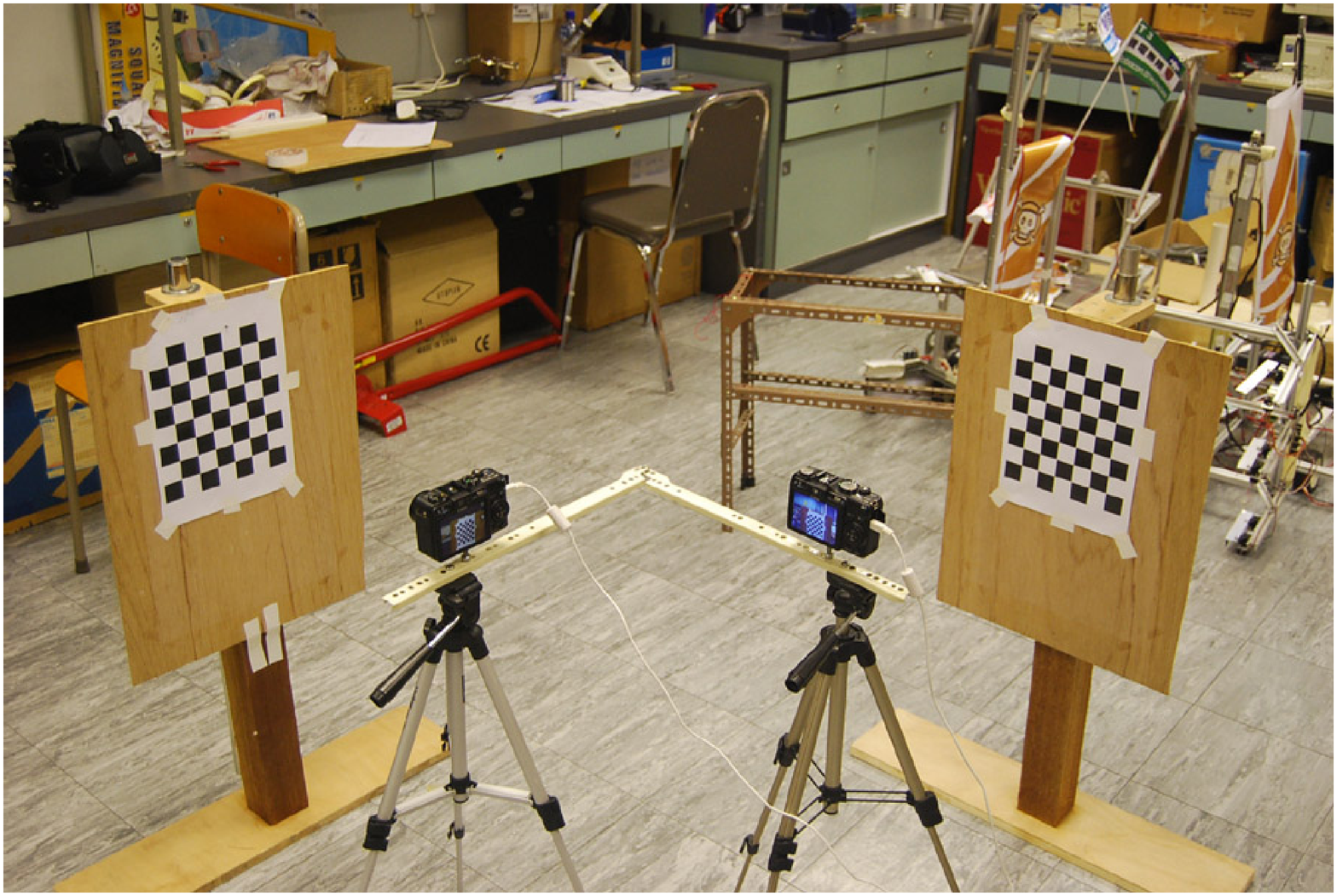}}\\
{(a)} & (b) \\
\includegraphics[width=1.1in]{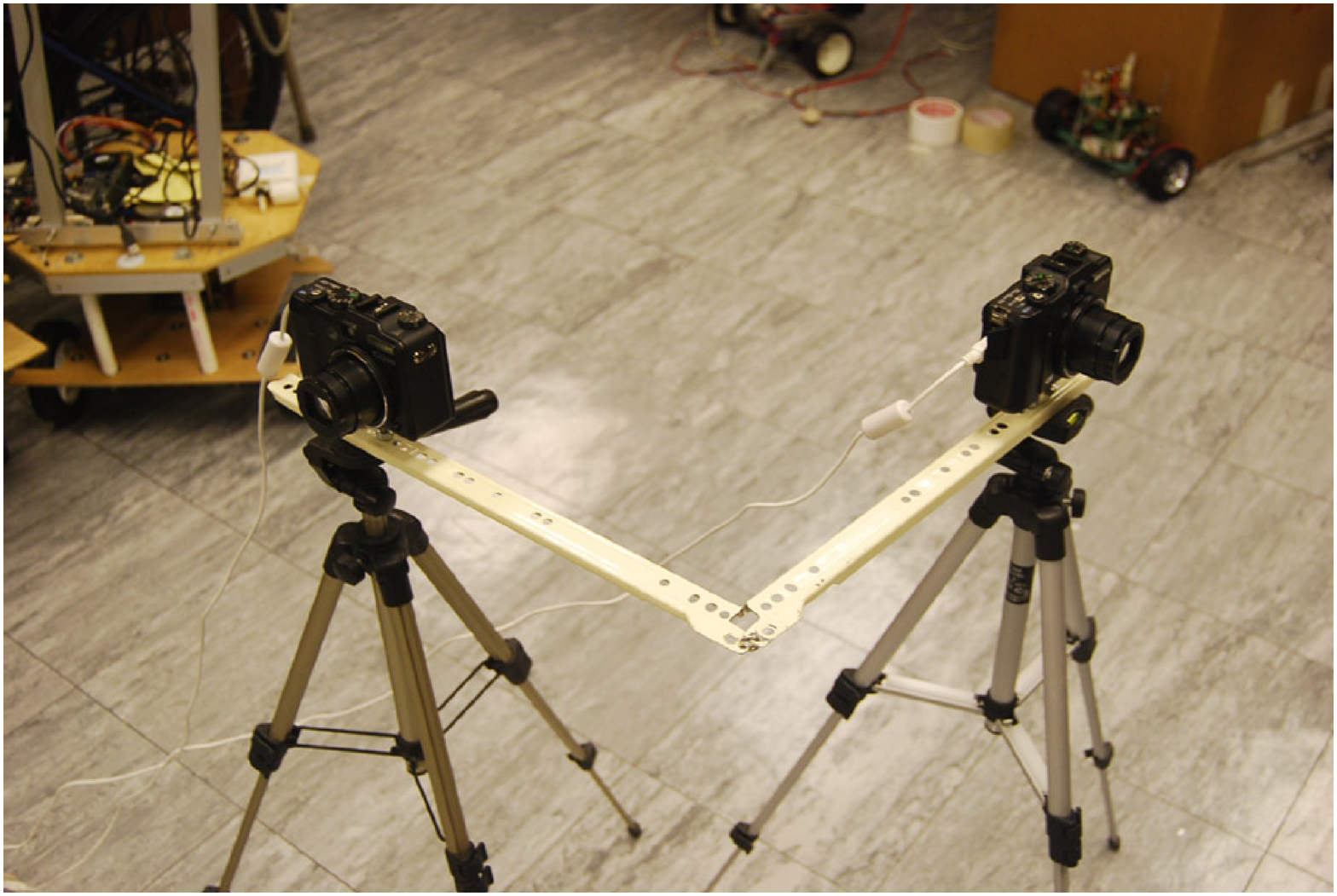}
& \includegraphics[width=1.1in]{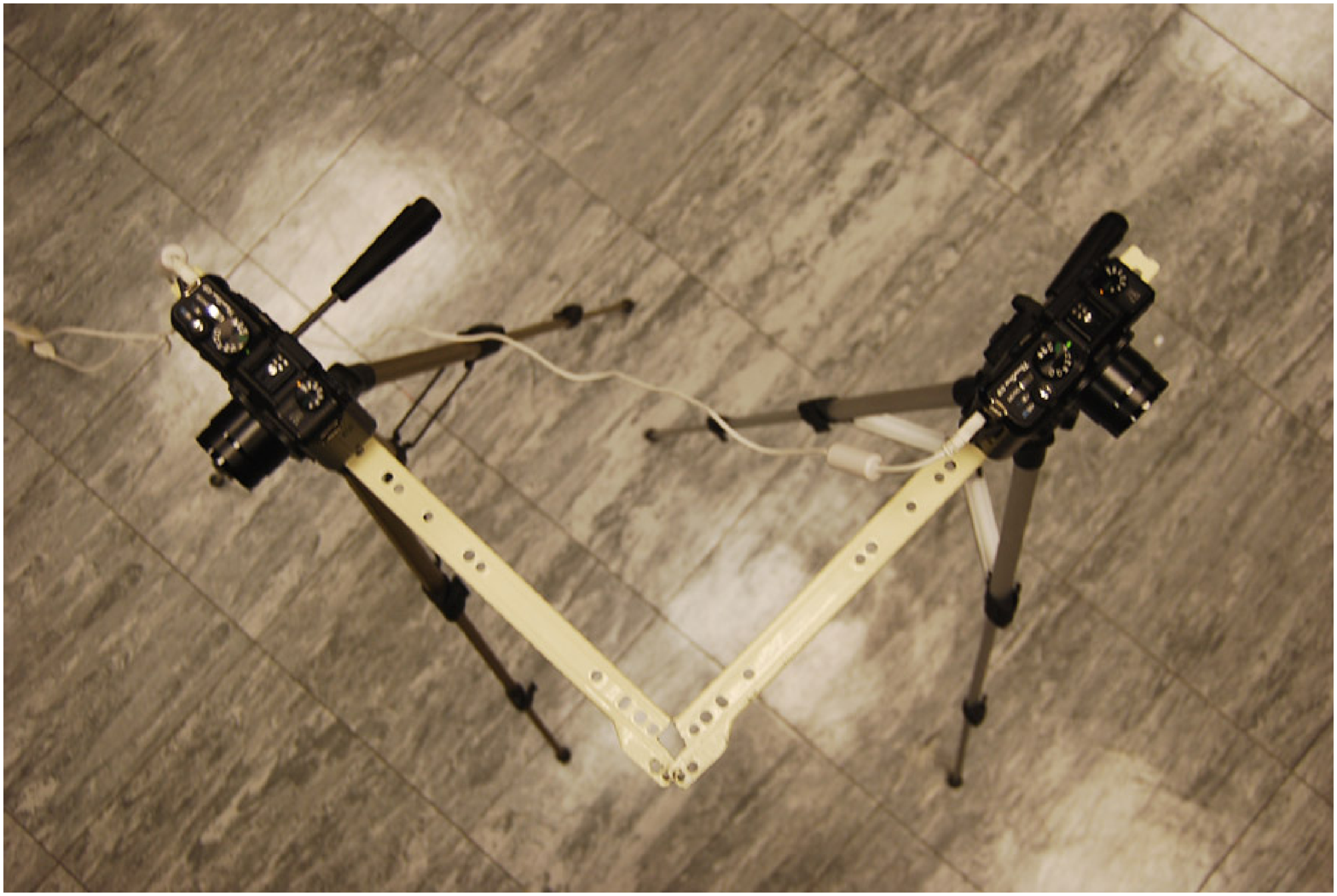}\\
(c)& (d)
\end{tabular}
\caption{The ACS with Two Cannon PowerShot G9 Used in the Real
Experiment. (a) The ACS Used in the Real Experiment. (b) The ACS and
two Checker Planes. (c) In the Front of the ACS. (d) On the Top of
the ACS.}\label{fig:non_overlap_cfg}
\end{figure}

\begin{figure}[tbh] \center
\small
 \begin{tabular}{cccc}
 \includegraphics[width=0.65 in]{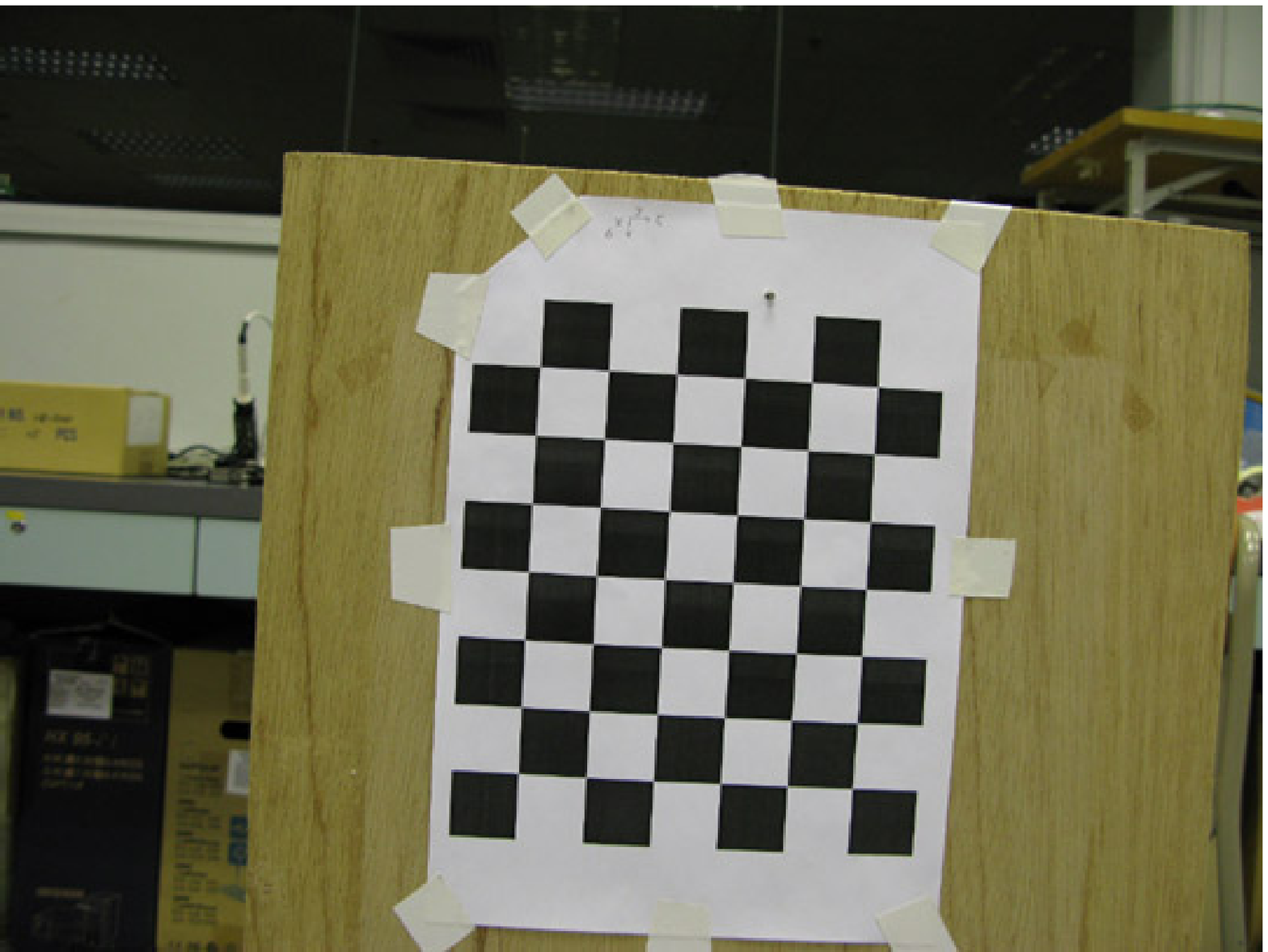}&
  \includegraphics[width=0.65 in]{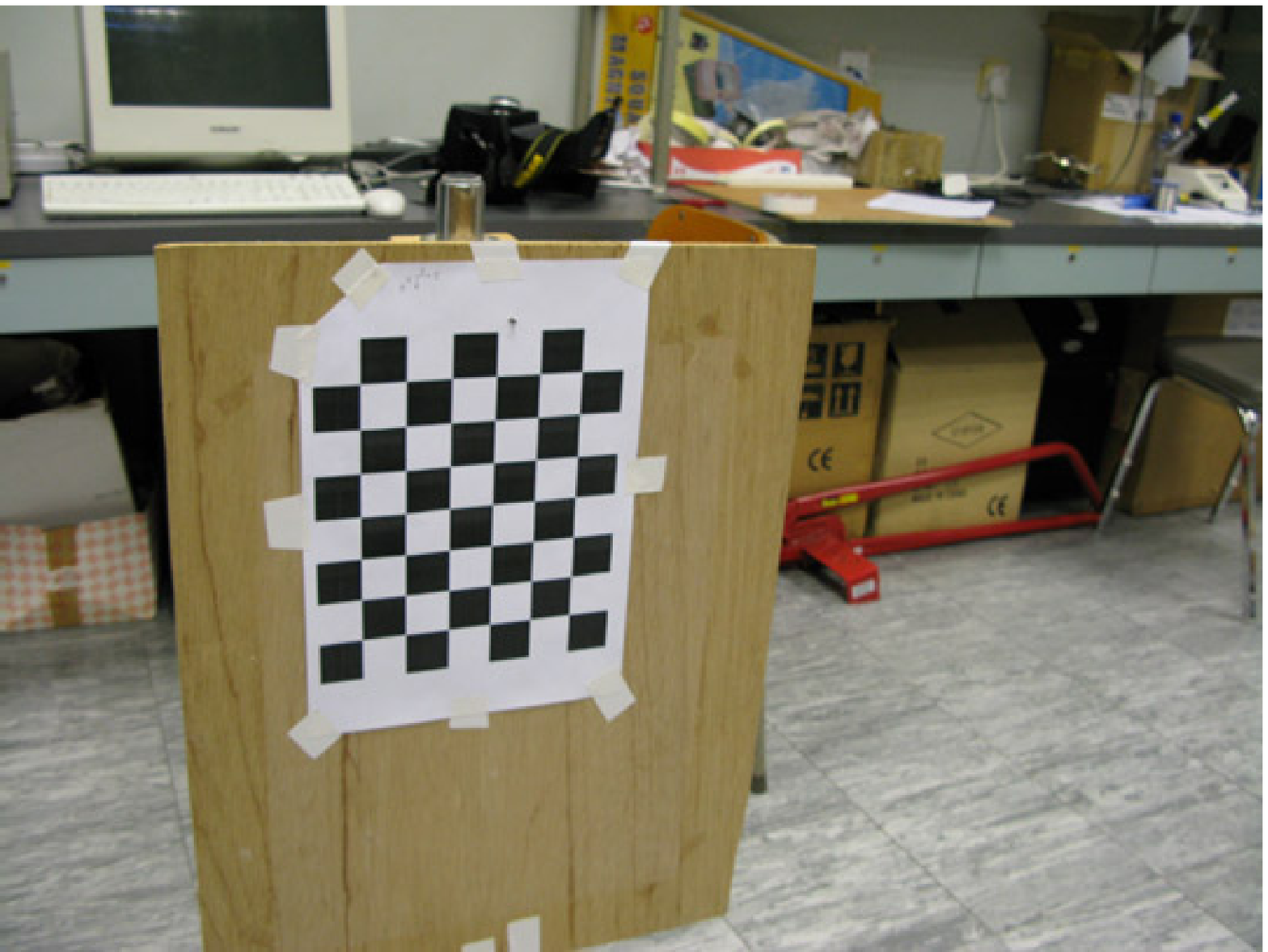}&
  \includegraphics[width=0.65 in]{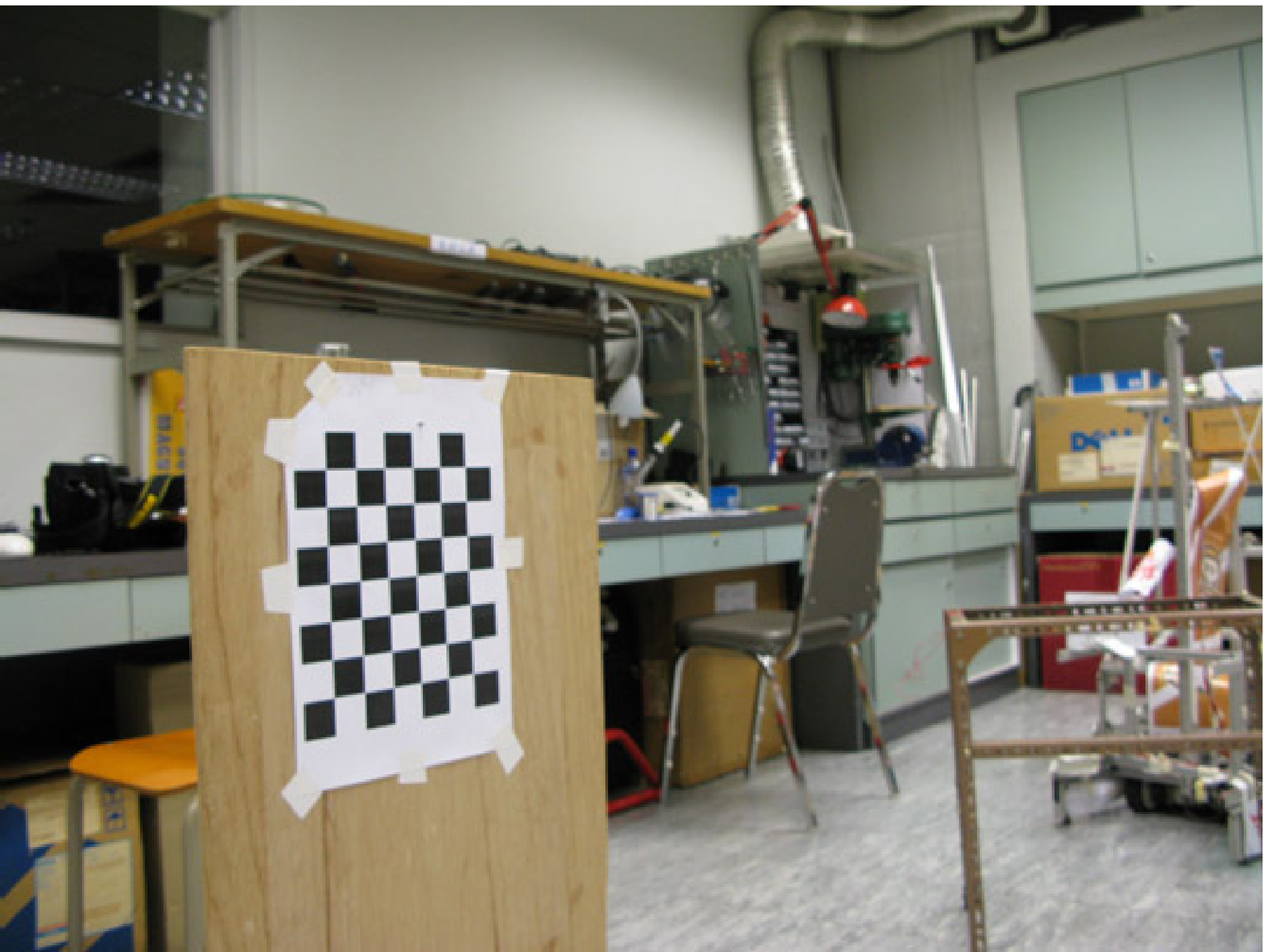}&
  \includegraphics[width=0.65 in]{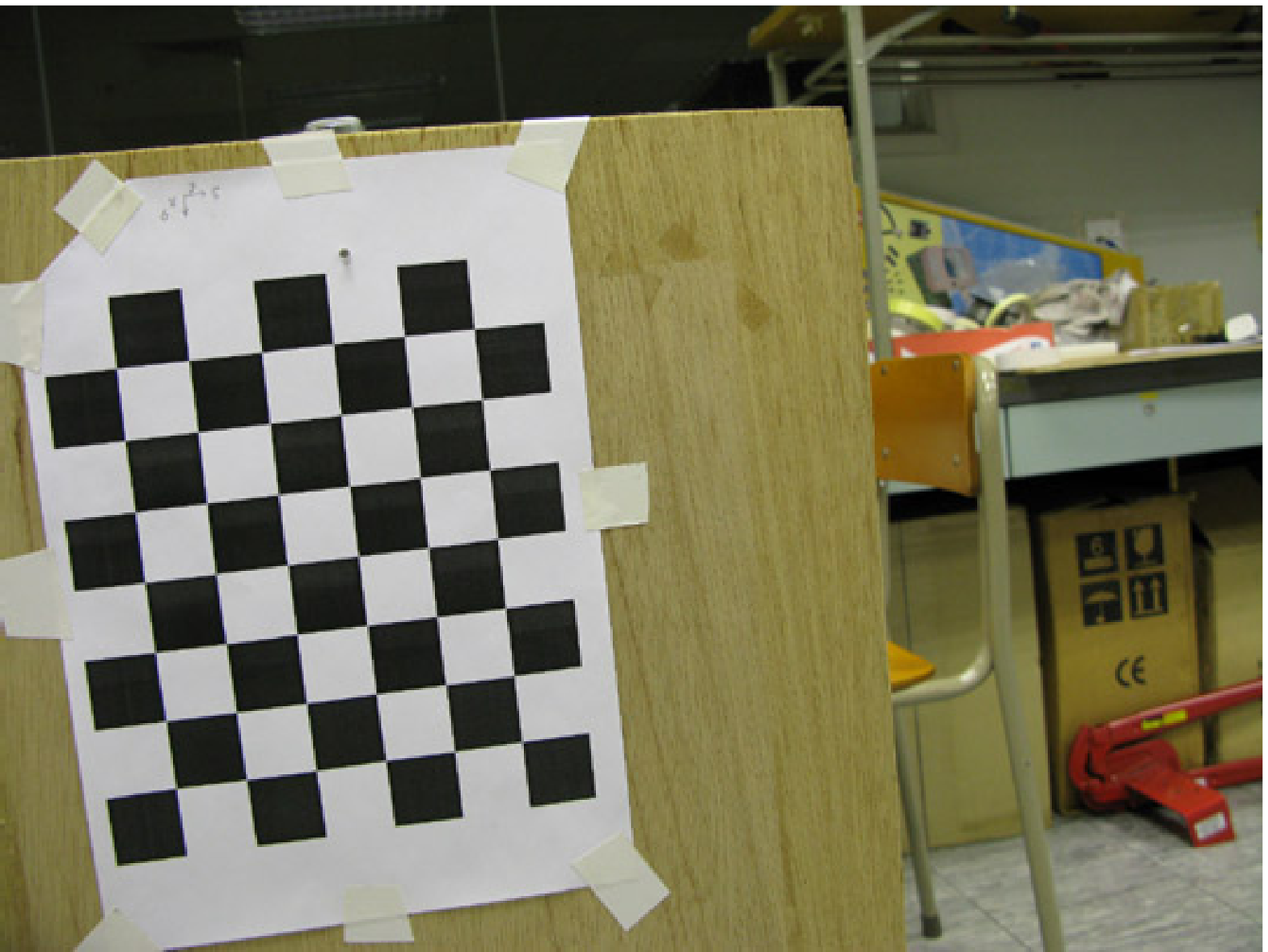}\\
  Img$_1$&Img$_6$&Img$_{12}$&Img$_{17}$\\
\end{tabular}\\
 (a) Images Recorded by Camera A\\
 \begin{tabular}{cccc}
  \includegraphics[width=0.65 in]{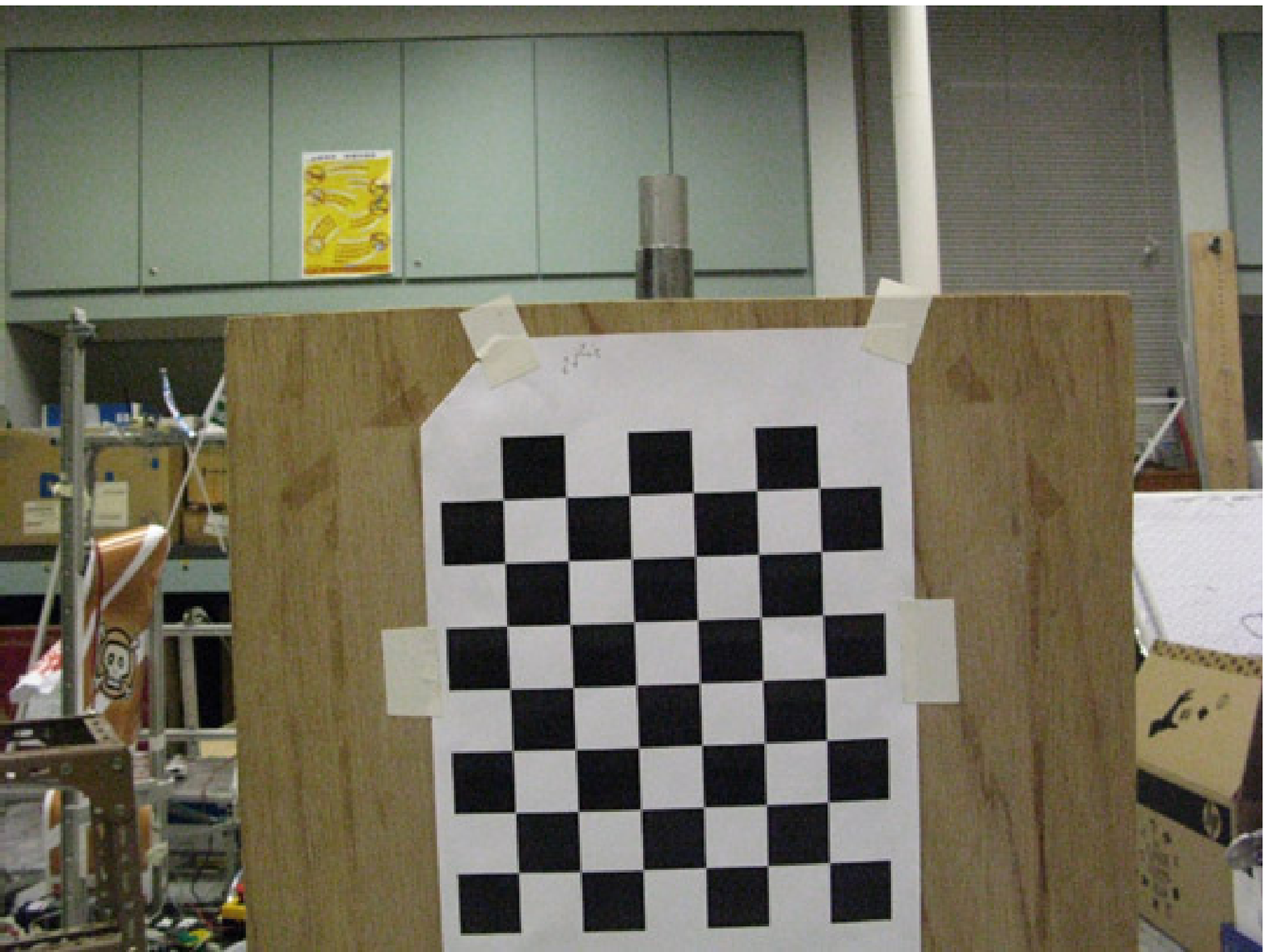}&
  \includegraphics[width=0.65 in]{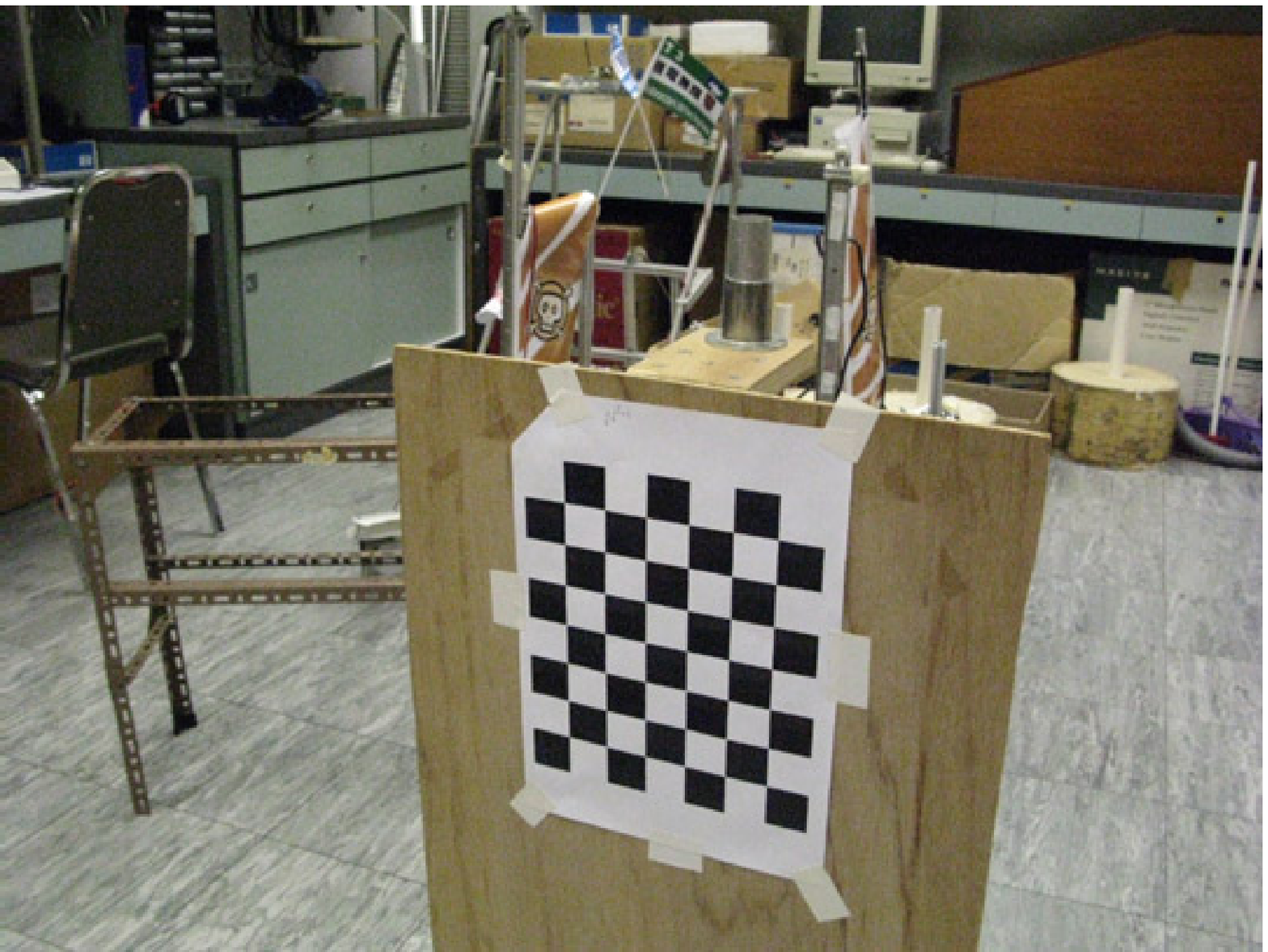}&
  \includegraphics[width=0.65 in]{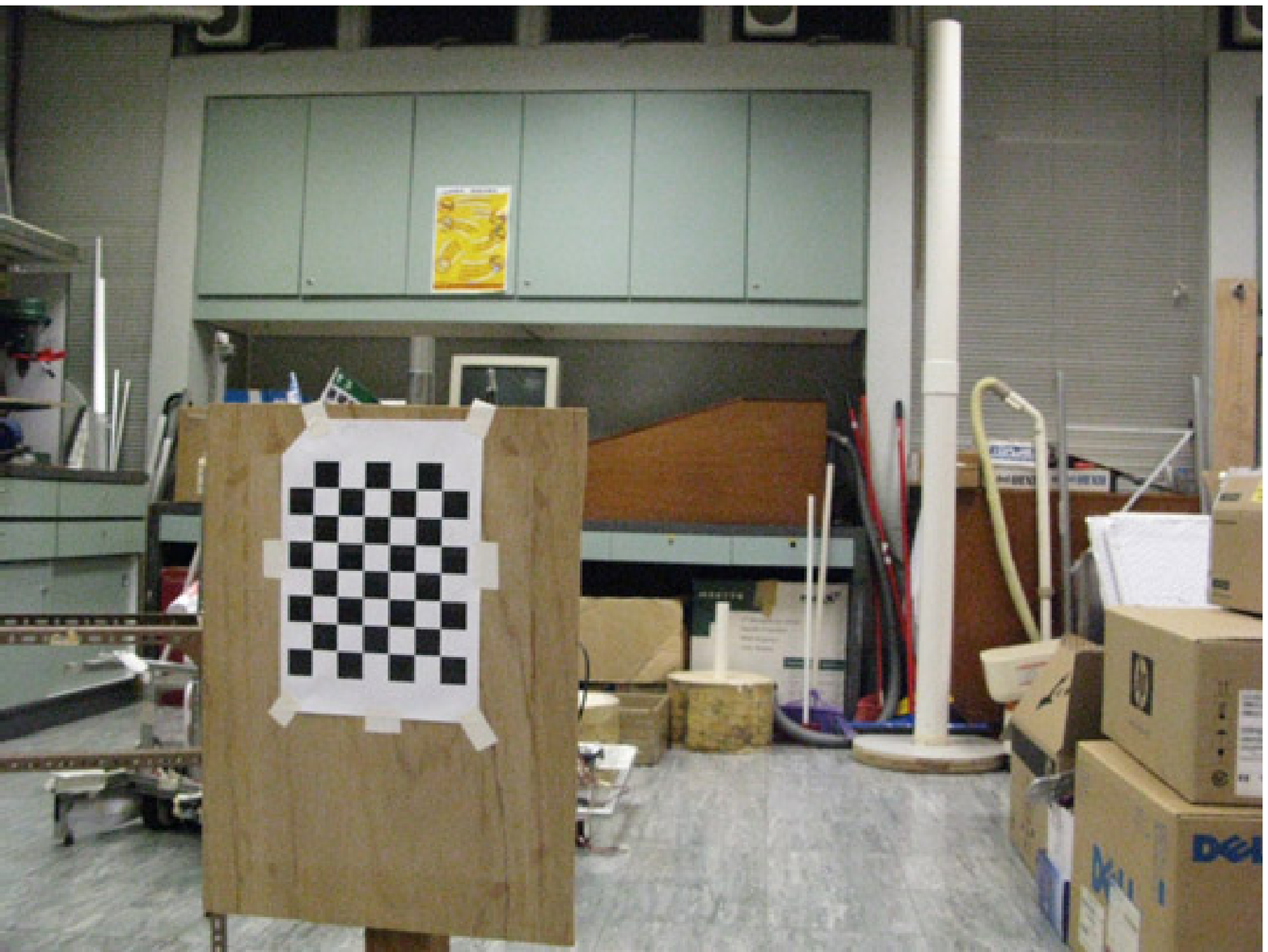}&
  \includegraphics[width=0.65 in]{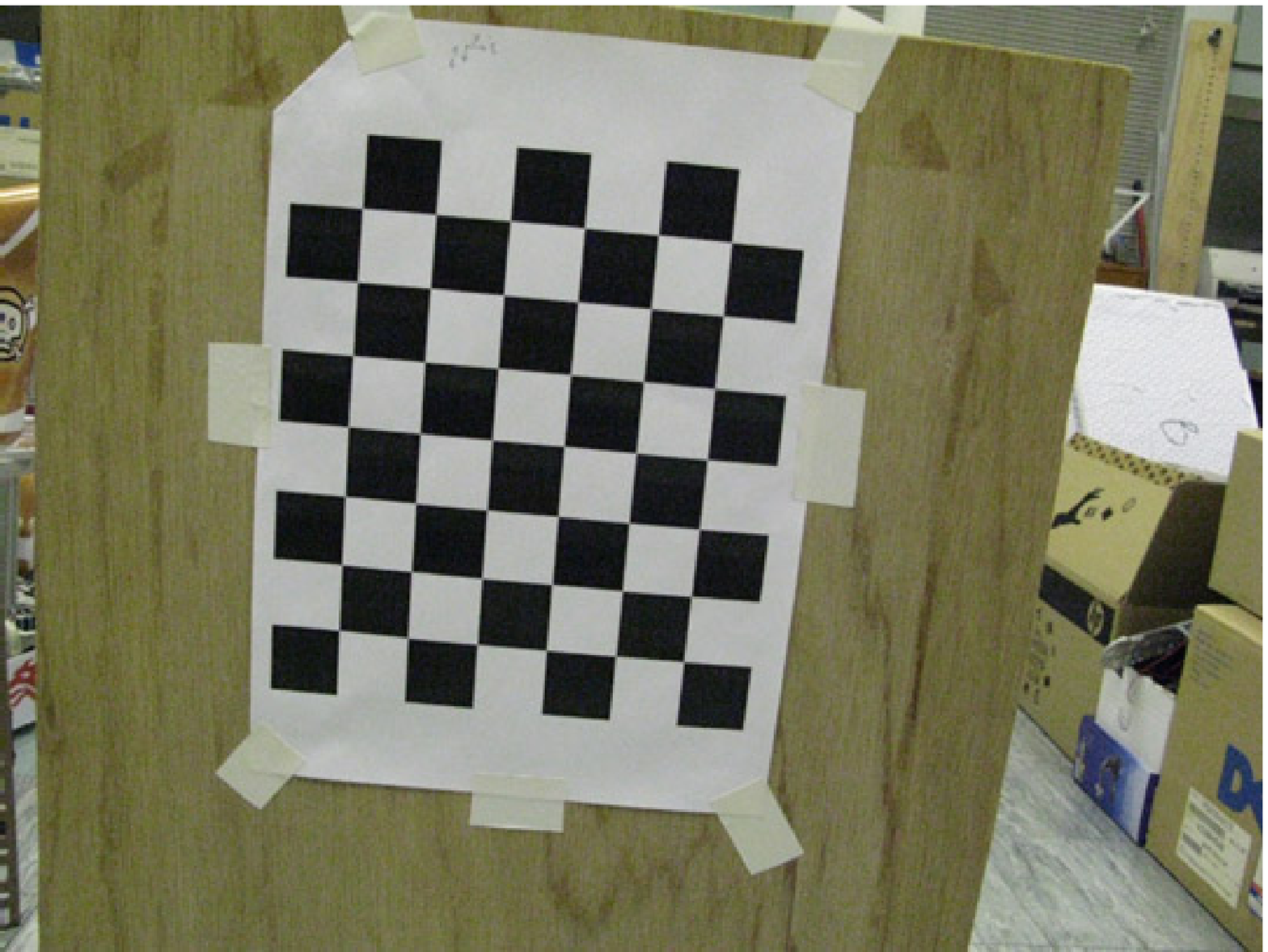}\\
  Img$_1$&Img$_6$&Img$_{12}$&Img$_{17}$\\
\end{tabular} \\
  (b) Images Recorded by Camera B
  \caption{Images Recorded by the ACS}\label{fig:cam}
\end{figure}

\begin{figure}
\centering
  \includegraphics[width=2.7in]{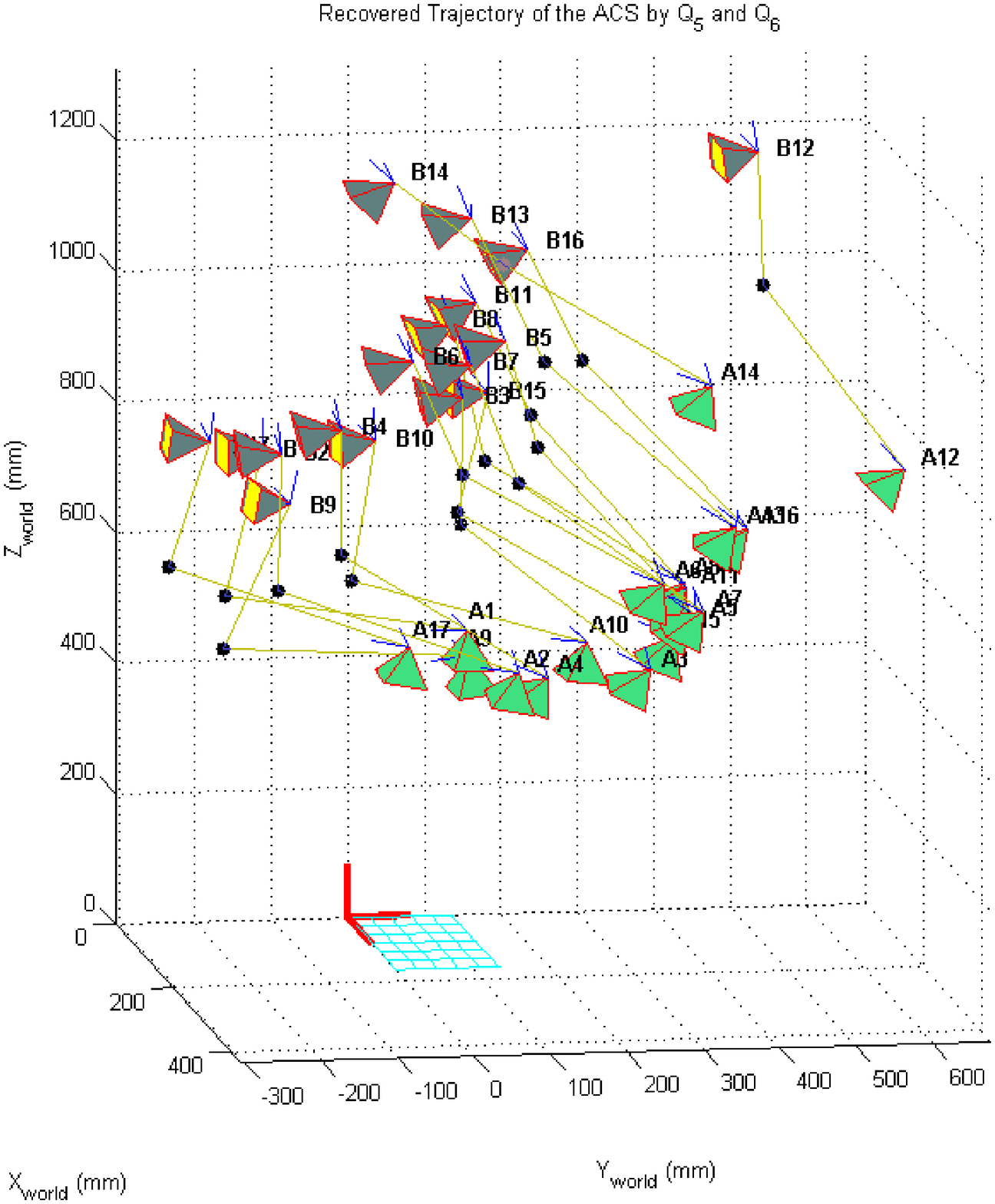}\\
  \caption{The Trajectory of the ACS Recovered from $Q_5$ and $Q_6$}\label{fig:traj_non_overlap}
\end{figure}

In the third
real experiment, firstly, we use the non-overlapping view ACS
calibration method to process the image sequences $Q_1$ and $Q_2$.
The joint pose (${\bar{O}}_A$) estimated by algorithm II is used as
the input for the relative pose calibration. Since there are
overlapping views between $Q_1$ and $Q_2$, we also calibrate the
relative pose between the two cameras by the feature correspondences
for comparison. The calibration result are listed in Table
\ref{tab:rt}. After the joint pose relative to each camera in the
ACS and relative pose between the cameras in the ACS are calibrated,
the trajectory of the ACS is recovered (see Figure
\ref{fig:traj_overlap}).
\begin{table}[h]
\caption{Result of Relative Pose Calibration}
\label{tab:rt}\centering
{\small III: our method. (see section
\ref{sec:alg_iii}) F: using feature correspondences.\\}
\begin{tabular}{|c|c|c|c|}
\hline{Algorithm}&\multicolumn{3}{c|}{Relative Rotation (Degree)}\\
\cline{2-4}
& Roll& Pitch& Yaw\\
\hline III  &17.7158 &  -11.3660 & -80.1913\\
\hline F &  17.5459 & -10.6024 & -78.9854\\
\hline
\hline{Algorithm}&\multicolumn{3}{c|}{Relative Translation (mm)}\\
\cline{2-4}
& $T_x$& $T_y$& $T_z$\\
\hline III  & 295.4183 & -232.4576  & 34.5004\\
\hline F &  294.0235 &-229.8369 &  28.9739\\
\hline
\end{tabular}
\end{table}
The proposed calibration method is also tested by non-overlapping
view image sequences. Figure \ref{fig:non_overlap_cfg} (b), (c), (d)
shows the configuration of the non-overlapping view ACS calibration
system in the real experiment. Two image sequences ($Q_5$ and $Q_6$)
are recorded, each sequence consists of $17$ images of size
$1600\times 1200$ pixels. There is no overlapping view between $Q_5$
and $Q_6$. Figure \ref{fig:cam} shows some samples of the recorded
images. We also manually measured the relative pose between the two
cameras for comparison. Since no feature correspondence can be used,
we only get a rough estimation by a ruler. The calibration results
are shown in Table \ref{tab:rt_non_overlap}. After the relative pose
between the cameras at the initial state is estimated, the
trajectory of the non-overlapping view ACS is recovered (see Figure
\ref{fig:traj_non_overlap}).
\begin{table}
\centering  \caption{Result of Relative Pose Calibration Using
Non-Overlapping View Image Sequences}\label{tab:rt_non_overlap}
{\small III: our method. (see section \ref{sec:alg_iii}) M: manual
measurement\\} \ \\ \
\begin{tabular}{|c|c|c|c|}
\hline{Algorithm}&\multicolumn{3}{c|}{Relative Rotation (Degree)}\\
\cline{2-4}
& Roll& Pitch& Yaw\\
\hline III  &1.3182 &  88.4530  &  0.7315\\
\hline M &  0 $\pm$ 5 & 90 $\pm$ 5   &  0 $\pm$ 5\\
\hline
\hline{Algorithm}&\multicolumn{3}{c|}{Relative Translation (mm)}\\
\cline{2-4}
& $T_x$& $T_y$& $T_z$\\
\hline III  & 291.3321&  -17.2837& -292.1382\\
\hline M &  290$\pm$20& 0 $\pm$ 20& 280 $\pm$20 \\
\hline
\end{tabular}
\end{table}
\subsection{Calibration of Relative Pose Between the Cameras in the
 Non-Overlapping View ACS with Unknown Scale Factors (Algorithm IV)}\label{sec:alg_iv}

The scale factor estimation algorithm is evaluated in the fourth
real experiment. The estimated translations from $Q_1$ and $Q_2$ are
multiplied by $0.8$ and $3.2$ respectively. In this case, if no
noise exists, the estimated relative scale factor ($\phi_{BA}$)
should be $4$. The estimated relative scale factor
($\hat{\phi}_{BA}$) in our experiment was $3.8919$. Table
\ref{tab:RelativeCalibration_scale} lists the corresponding results,
in which the estimated relative translations are divided by $3.2$,
so that they can be easily compared with the estimated relative
translations in Table \ref{tab:rt}. The experiment showed that our
algorithms can estimate the relative scale factor and find the
extrinsic parameters correctly. In order to test the stability of
the scale factor estimation algorithm, the estimated translations
from $Q_5$ and $Q_6$ are multiplied by $0.8$ and $3.2$ respectively.
$100$ tests are performed. In each test, $22$ images are randomly
selected as section \ref{sec:non_overlap_exp}. The Mean and STD of
the calibration results is listed in Table
\ref{tab:rt_non_overlap_static_scale}. The results are good.

\begin{table}[htbp]
\centering  \caption{Result of Relative Pose Calibration with
Unknown Scale Factors (0.8 in $Q_1$ and 3.2 in $Q_2$
)}\label{tab:RelativeCalibration_scale}
{\flushleft
\small IV: our scale factor estimation method. (see section
\ref{sec:alg_iv}) F: using feature correspondences.}

\begin{tabular}{|c|c|c|c|}
\hline{Algorithm}&\multicolumn{3}{c|}{Relative Rotation (Degree)}\\
\cline{2-4}
& Roll& Pitch& Yaw\\
\hline IV  & 17.4883 & -10.5185 & -79.2551\\
\hline F &  17.5459 & -10.6024 & -78.9854\\
\hline
\hline{Algorithm}&\multicolumn{3}{c|}{Relative Translation (mm)}\\
\cline{2-4}

& $T_x$& $T_y$& $T_z$\\
\hline IV  & 295.9218 & -220.6804 & 11.5566 \\
\hline F &  294.0235 &-229.8369 & 28.9739 \\
\hline
\end{tabular}
\end{table}

\begin{table}
\centering  \caption{Mean and STD of the Relative Pose Calibration
Using Non-Overlapping View Image Sequences with Unknown Scale
Factors. ($Q_5$ and $Q_6$) (see section \ref{sec:alg_iv})}
\label{tab:rt_non_overlap_static_scale}
\begin{tabular}{|c|c|c|c|}
\hline{Algorithm}&\multicolumn{3}{c|}{Relative Rotation (Degree)}\\
\cline{2-4}
IV& Roll& Pitch& Yaw\\
\hline Mean  & -4.4275 & 38.9820 & -14.3572\\
\hline STD &  0.4304 & 0.2639  & 0.5774 \\
\hline
\hline{Algorithm}&\multicolumn{3}{c|}{Relative Translation (mm)}\\
\cline{2-4}
IV& $T_x$& $T_y$& $T_z$\\
\hline Mean  & 489.2497 & -56.0786 & -165.7425\\
\hline STD &6.2496  & 3.1070  & 3.7616\\
\hline
\hline{Algorithm}&\multicolumn{3}{c|}{Relative Scale Factor}\\
\cline{2-4}
IV& \multicolumn{3}{c|}{$\phi_{BA}$}\\
\hline Mean &\multicolumn{3}{c|}{3.9531}\\
\hline STD & \multicolumn{3}{c|}{0.0159}\\
\hline
\end{tabular}
\end{table}

\section{Conclusion}\label{sec:conclusion}
In this paper, an ACS calibration method is developed. Both the
simulation and real experiment show that the pose of the joint in an
ACS can be estimated robustly. When there is no overlapping view
between the cameras in an ACS, the joint pose and the relative pose
between the cameras can also be calculated. The trajectory of an ACS
can be recovered after the ACS is calibrated. The proposed
calibration method requires only the image sequences recorded by the
cameras in the ACS. A scale factor estimation algorithm is proposed
to deal with unknown scale factors in the estimated translation
information of the cameras in an ACS. In the real experiment, the
intrinsic and extrinsic parameters of the ACS are calibrated using
the same image sequences simultaneously.

Since we still cannot find any former study of the ACS calibration
in the literature. We apologize for having no comparison with former
ACS calibration method.

Our future plan may focus on using an ACS attached on different
parts of human body to track the motion of the human. We foresee
that if calibration of articulated cameras become a simple routine,
researchers will find many novel and interesting applications for
such a camera system.

\end{document}